\documentclass[journal]{IEEEtran}
\usepackage{amsmath,amsfonts}
\usepackage{algorithm}
\usepackage{algpseudocode}
\usepackage{array}
\usepackage[caption=false,font=normalsize,labelfont=sf,textfont=sf]{subfig}
\usepackage{textcomp}
\usepackage{stfloats}
\usepackage{url}
\usepackage{verbatim}
\usepackage{graphicx}
\usepackage{cite}
\usepackage{booktabs}
\usepackage[table]{xcolor}
\usepackage{multirow}
\usepackage{bbding}
\usepackage{pifont}

\usepackage[colorlinks=true,citecolor=blue,linkcolor=blue,urlcolor=black]{hyperref}

\begin{document}

\title{Multi-Modal~Object~Re-Identification~with~Prompt-S6~and~Semantic-Aware~Knowledge~Guidance}

\author{Weixiang Zhou, Jiabei Zuo, Yuhao Wang, Cong Wang,
        Huchuan Lu,~\IEEEmembership{Fellow,~IEEE,}
        and Zhixun Su,~\IEEEmembership{Member,~IEEE}

\thanks{ This work was supported in part by the National Natural Science Foundation of China under Grant 62476041. \textit{(Weixiang Zhou and Jiabei Zuo contributed equally to this work.) (Corresponding author: Zhixun Su and Cong Wang.)}}

\thanks{
Weixiang Zhou and Zhixun Su are with the School of Mathematical Sciences and the Key Laboratory for Computational Mathematics and Data Intelligence of Liaoning Province, Dalian University
of Technology, Dalian, 116024, China (e-mail: s20201162006@mail.dlut.edu.cn; zxsu@dlut.edu.cn).}

\thanks{Jiabei Zuo is with the School of Computer Science and Technology, Dalian University of Technology, Dalian, 116024, China (e-mail: 2214944912@mail.dlut.edu.cn).}

\thanks{Yuhao Wang and Huchuan Lu are with the School of Future Technology, School of Artificial Intelligence, Dalian University of Technology, Dalian, 116024, China (e-mail: 924973292@mail.dlut.edu.cn; lhchuan@dlut.edu.cn).}

\thanks{Cong Wang is with the Department of Radiology and Biomedical Imaging, University of California, San Francisco, 94107, USA (e-mail: supercong94@gmail.com).

}
}

\markboth{}{}
\maketitle

\begin{abstract}
Multi-modal object Re-Identification (ReID) aims to retrieve specific objects by integrating complementary information from multiple modalities. However, existing multi-modal ReID methods do not effectively address background interference suppression or achieve tri-modal alignment, instead focusing on pairwise feature fusion. Moreover, many current aggregation approaches suffer from high computational complexity.
To address these limitations, we propose PRISM, a novel multi-modal ReID framework built upon Prompt-S6 (PS6) and semantic-aware knowledge guidance. PS6 maintains the linear complexity and strong sequence modeling capability of Mamba while enabling efficient cross-modal interaction. Leveraging these advantages, we design two key components: Semantic-Driven Token Pruning (SDTP) and Progressive Fusion Network (PFN).
Parsing semantic priors from the segmentation foundation models, the SDTP then leverages these priors and applies dynamic token pruning to suppress background noise and refine feature representations. The PFN progressively aggregates multi-modal features to achieve tri-modal alignment and fully exploit modality complementarity.
With the proposed modules, PRISM generates more robust multi-modal representations under complex scenarios.
Extensive experiments on four multi-modal object ReID benchmarks demonstrate the effectiveness and efficiency of our approach.
The source code is available at \url{https://github.com/zw-absin/PRISM}.
\end{abstract}


\begin{figure*}
    \centering
    \includegraphics[width=1.0\textwidth]{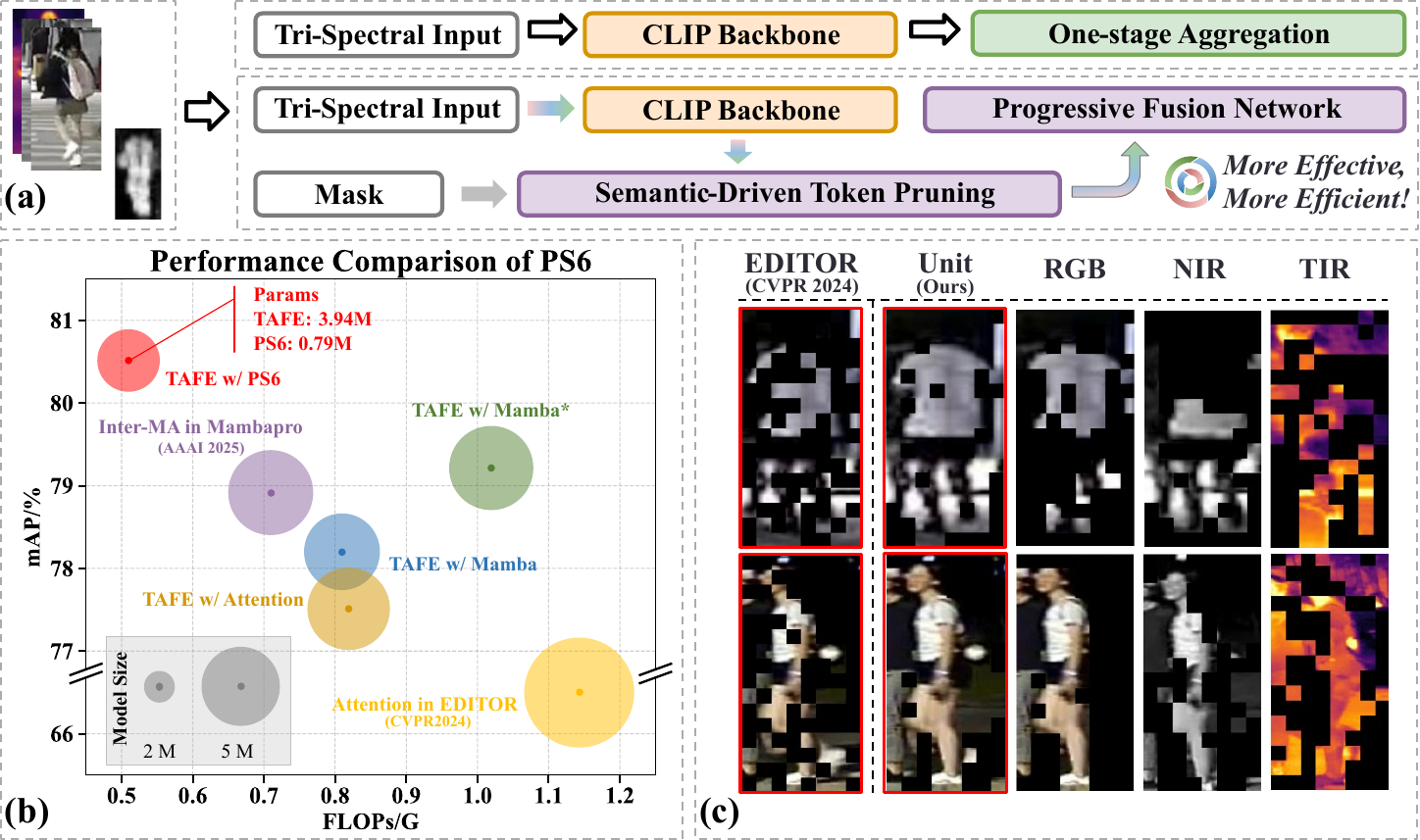}
    \caption{\textbf{(a)} Our method addresses two key limitations: ignoring background information and struggling with tri-modal alignment. These are addressed through semantic-driven token pruning and a progressive fusion network.
    \textbf{(b)} Enabled by PS6, our framework achieves stronger cross-modal interactions and improved performance. 
    \textbf{(c)} Token selection visualization demonstrates that our method retains more informative regions.}
    \label{fig1}
\end{figure*}

\section{Introduction}\label{sec:Introduction}
\IEEEPARstart{O}{bject} Re-Identification (ReID) aims to match instances of the same object across non-overlapping camera views. Owing to its broad applicability in surveillance, autonomous driving, and smart city systems, single-modal ReID based on RGB imagery has seen substantial progress in recent years~\cite{wang2020receptive,liu2023deeply,wang2024other,liu2024video,liu2021watching,zhang2021hat,yu2024tf}.
However, RGB-based methods are highly sensitive to environmental variations such as low illumination, glare, and occlusion. These challenges often lead to significant performance degradation in adverse conditions~\cite{wang2020joint,wang2025ultra,sun2025adapting,wang2026neural,wang2025deep}, limiting their robustness and generalization in real-world deployments~\cite{wang2024top}.
To address these limitations, multi-modal ReID has emerged as a promising alternative by exploiting complementary cues from heterogeneous sensors. This approach integrates information from multiple modalities, such as multi-spectral images, depth maps, and textual descriptions, to construct more discriminative and resilient feature representations. As a result, multi-modal ReID achieves superior performance in degraded scenarios where RGB-only systems fail~\cite{shi2024learning,yang2024shallow,Yang_2023_ICCV,li2025video,tao2026spatial,zhou2026hierarchical,leng2026dynamic}.

However, existing multi-modal methods often overlook object-background separation by extracting features directly from raw multi-spectral inputs. 
Under realistic imaging conditions, such inputs are frequently contaminated by noisy and cluttered backgrounds, which degrade model robustness~\cite{ye2021deep} and impede accurate cross-modal feature alignment in ReID~\cite{zhang2024magic}.
To mitigate this, some works incorporate semantic masks to isolate foreground regions~\cite{song2018mask,qi2019mask, zhang2023multi,cui2024profd,li2025shape}, while others exploit pre-trained vision-language models like CLIP~\cite{radford2021learning} for enhanced semantic understanding through prompt learning and cross-modal alignment~\cite{wei2025multiple,yan2023clip}. 
Yet, these methods are primarily designed for single-modal settings and struggle to preserve discriminative semantic parts under challenging conditions such as occlusion, background clutter, and cross-modal misalignment.
More recently, token selection mechanisms have been proposed to focus on salient regions. 
Zhang et al.~\cite{zhang2024magic} retain tokens with high attention weights. 
While improving feature concentration, such strategies risk discarding potentially useful contextual information and may retain background-correlated tokens due to attention bias.
To fully aggregate tri-modal features, Wang et al.~\cite{wang2024top} utilize Multi-Head Cross-Attention (MHCA)~\cite{dosovitskiy2020image} to enable each modality's class token to perceive information from other modalities. 
Although this facilitates pairwise interactions, it limits the ability to achieve comprehensive tri-modal alignment due to the isolation of the third modality.
These limitations highlight the need for a more effective mechanism that simultaneously ensures semantic fidelity, maximizes information utilization, and achieves comprehensive tri-modal alignment.

To address the issues of imprecise token selection and insufficient tri-modal alignment in existing methods, we propose PRISM, a novel framework for robust multi-modal object ReID.
As shown in Fig.~\ref{fig1}(a), PRISM first extracts multi-spectral features from RGB, Near-Infrared (NIR), and Thermal-Infrared (TIR) images using a shared vision encoder. 
Concurrently, a pre-trained mask extractor generates semantic masks, which provide spatially aligned guidance to distinguish foreground from background.
To enhance semantic fidelity and suppress background interference, we introduce Semantic-Driven Token Pruning (SDTP), comprising two components: 
(i)~Target-Aware Feature Enhancement (TAFE), which leverages Prompt-S6 (PS6) for efficient fine-grained fusion between semantic masks and multi-spectral features (Fig.~\ref{fig1}(b)); and (ii)~Tail Drop Module (TDM), which employs a cross-modal consensus mechanism to identify and suppress tokens likely associated with background or non-discriminative regions, as shown in Fig.~\ref{fig1}(c). 
By preserving the majority of foreground-aligned tokens, TDM achieves a favorable balance between noise suppression and feature completeness.
Further, we propose the Progressive Fusion Network (PFN), which integrates intra-modal modeling, inter-modal interaction, and tri-modal aggregation in a three-stage strategy. 
This design progressively refines feature representations and enables comprehensive alignment across all three modalities.
Both TAFE and PFN are built upon Prompt-S6 (PS6), a lightweight module grounded in the theoretical framework of State Space Models (SSMs)~\cite{gu2021efficiently}. Unlike attention-based approaches, PS6 enables efficient and simultaneous tri-modal interaction.
Through these components, PRISM effectively leverages semantic structure to guide feature refinement and achieves robust cross-modal alignment.
Extensive experiments on four multi-modal benchmark datasets validate the effectiveness of our proposed method. To summarize, our main contributions are as follows:

\begin{itemize}
    \item We propose Prompt-S6 (PS6), a lightweight cross-modal interaction module grounded in State Space Models (SSMs). PS6 enables efficient, simultaneous tri-modal interaction with linear complexity and fewer parameters, avoiding the quadratic overhead of attention mechanisms.

    \item We introduce Semantic-Driven Token Pruning (SDTP), which leverages semantic masks to enhance foreground features and suppress background-associated tokens through a cross-modal consensus mechanism, improving feature selectivity.

    \item We design the Progressive Fusion Network (PFN), a three-stage fusion pipeline that progressively models intra-modal dynamics, inter-modal interactions, and tri-modal alignment for comprehensive feature integration.

    \item Based on PS6, SDTP, and PFN, we present PRISM, a novel framework for multi-modal object ReID that achieves robust semantic alignment and efficient feature learning across RGB, NIR, and TIR modalities.
\end{itemize}

\section{Related Works}\label{sec:RelatedWorks}
We review related works in three aspects: single-modal object Re-Identification (ReID), multi-modal object ReID, and State Space Models (SSMs) for visual sequence modeling. 

\subsection{Single-Modal Object Re-Identification}
Due to the demands of real-world applications, single-modal object ReID has witnessed rapid development. 
This line of work focuses on extracting discriminative features from a single spectral modality, primarily including RGB, Near-Infrared (NIR), Thermal-Infrared (TIR), and depth images, among which RGB is the most prevalent. 
State-of-the-art methods in this domain are largely built upon Convolutional Neural Networks (CNNs) or Transformers~\cite{vaswani2017attention}. 
For example, OSNet~\cite{zhou2019omni} introduces a multi-stream residual block to achieve holistic multi-scale feature learning. 
He et al.~\cite{he2021transreid} propose the first pure Transformer-based architecture for ReID, aiming to preserve fine-grained details by eliminating convolutional operations and spatial downsampling.
In addition, several approaches incorporate semantic priors to enhance the model's semantic awareness. 
Song et al.~\cite{song2018mask} and Qi et al.~\cite{qi2019mask} employ binary foreground masks to separate and model the target object from the background. 
Zhu et al.~\cite{zhu2020identity} leverage part-level semantic segmentation masks to distinguish different body regions. 
Cui et al.~\cite{cui2024profd} and Somers et al.~\cite{somers2024keypoint} integrate human keypoint detection to guide feature learning. 
However, like most single-modal methods, these approaches are inherently limited by the characteristics of a single modality and exhibit significant performance degradation under severe conditions, such as low illumination, occlusion, or adverse weather. 
This performance gap is partly attributed to the reduced effectiveness of their semantic generation and fusion mechanisms under such conditions.
Many rely on fine-grained semantic priors (e.g., masks or keypoints) that are typically derived from clean, well-lit inputs~\cite{zhu2020identity,somers2024keypoint}. 
Others employ binary masking strategies that rigidly distinguish foreground from background~\cite{song2018mask, qi2019mask}, but fail to model nuanced semantic interactions or adapt to degraded visual conditions. 
When these fragile signals are processed through simplistic fusion pathways, the resulting representations become unreliable, further degrading overall performance.

To address these limitations, we introduce a robust semantic guidance framework that overcomes the fragility of traditional mask-based priors under degraded conditions. 
Instead of relying on binary masks or part annotations generated from clean inputs, our approach leverages prompt-driven soft masks to establish fine-grained, token-level interaction between semantic cues and multi-spectral features. 
This enables selective refinement of critical semantic regions, ensuring that guidance remains effective even when appearance is severely degraded.

\subsection{Multi-Modal Object Re-Identification}
Compared to single-modal ReID, multi-modal ReID enhances robustness in challenging environments by fusing complementary information from multiple sensing modalities~\cite{niu2020improving,zheng2021robust, wang2022interact, li2020multi, he2023graph, pan2023progressively, dong2025escaping, li2022visible, zheng2022visible}. 
Recent advances primarily focus on effective cross-modal modeling, with a growing trend toward adopting Vision Transformer (ViT)~\cite{dosovitskiy2020image} in multi-modal ReID due to its ability to model long-range dependencies and support flexible token-based representation~\cite{pan2023progressively, crawford2023unicat, zhang2024magic, wang2024top, wang2024other}.
To enhance inter-modal interaction, several methods design specialized architectures for token-level fusion. 
For instance, Wang et al.~\cite{wang2024top} introduce a cyclic token permutation framework to enable cross-modal information exchange, while Wang et al.~\cite{wang2024heterogeneous} propose a test-time training strategy that adapts to modality-specific variations during inference.
Concurrently, pre-trained multi-modal models such as CLIP~\cite{radford2021learning} have been leveraged as powerful feature extractors in ReID frameworks.
Building upon this, several approaches introduce learnable cross-modal prompts to enhance discriminative capability. Wang et al.~\cite{wang2025mambapro} employ modality-specific prompts and adapters with a frozen ViT encoder and leverage Mamba’s efficiency to significantly reduce FLOPs. In contrast, Zhang et al.~\cite{zhang2025prompt} keep the ViT backbone trainable and utilize shared cross-modal prompts for effective modality alignment and fusion, achieving superior generalization and notable performance gains on large-scale vehicle datasets. Other works explore Mixture-of-Experts (MoE) architectures. Feng et al.~\cite{feng2025multi} jointly model modality-shared and modality-specific features through pixel-level cross-modal interaction and sparse expert routing, attaining competitive accuracy with lower computational cost.
Wang et al.~\cite{wang2025decoupled} utilize dynamic attention mechanisms to handle decoupled multi-modal features.
However, these methods often struggle to effectively suppress background noise due to insufficient foreground-background discrimination, and their deployment efficiency is hampered by excessive prompt usage, stacked Mamba layers, or complex MoE routing mechanisms.
To address these limitations, we design an efficient fine-grained interaction module that integrates semantic information with multi-spectral features, thereby obtaining more robust and discriminative representations.

\subsection{Visual State Space Models in Re-Identification}
State Space Models (SSMs)~\cite{gu2021efficiently,gu2021combining,smith2022simplified} enable effective management of long sequence data with linear complexity, efficiently handling long-range dependencies. 
Among these, Mamba~\cite{gu2023mamba} has become a focal point of research due to its high efficiency and scalability in processing long sequences~\cite{wan2024sigma, xu2024polyp}. 
It improves upon SSMs through selective scan algorithms and hardware-aware optimizations, enabling linear complexity for long sequence data processing and outperforming Transformers~\cite{vaswani2017attention}. 
In the visual domain, Mamba demonstrates strong performance in tasks such as image classification, image processing, and multi-modal tasks~\cite{rahman2024mamba,zhu2024vision,liu2024vmamba,he2025pan}. 
Given their strong performance, SSMs have been introduced into ReID tasks in recent years. 
Geng et al.~\cite{geng2024remamba} propose a hybrid CNN-Mamba framework for local and global feature learning. 
Yu et al.~\cite{yu2025climb} apply Mamba for video processing, capturing critical spatiotemporal information. 
Zhang et al.~\cite{zhang2024mambareid} use multi-stage VMamba~\cite{liu2024vmamba} as a feature extractor. 
Wang et al.~\cite{wang2025mambapro} model intra-modal and inter-modal interactions using Mamba.
However, these methods mainly treat SSMs as substitutes for CNNs or Transformers in feature extraction or aggregation, without explicit modality alignment. 
To better address tri-modal alignment with reduced computational overhead, we introduce a fine-grained semantic interaction module and a progressive fusion module based on PS6, enabling efficient multi-modal feature aggregation.

\begin{figure*}
    \centering
    \includegraphics[width=\textwidth]{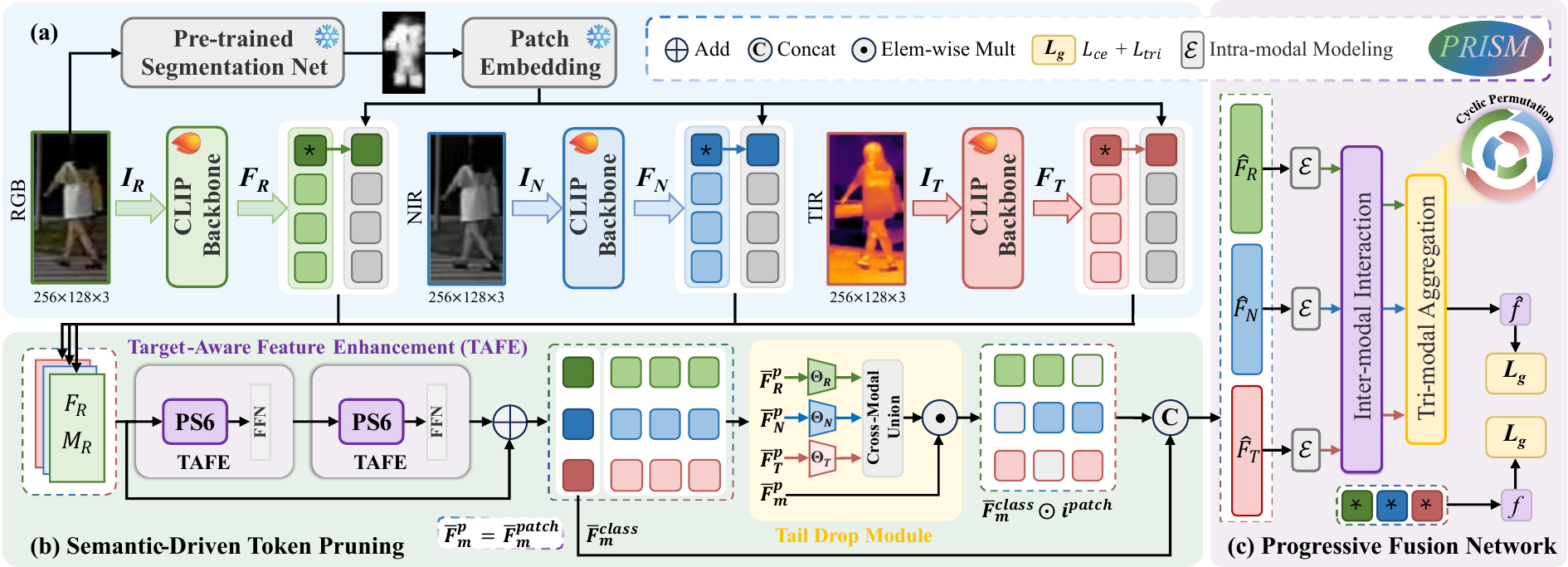}
    \caption{Overview of the proposed PRISM framework. 
    (a) Semantic masks and multi-spectral features are extracted using a segmentation network and vision encoder. 
    (b) To suppress background interference and exploit sequential information in a semantic-guided manner, the semantic masks and multi-spectral features interact in a fine-grained way, enabling knowledge distillation and pruning of irrelevant tokens.
    (c) Finally, for tri-modal alignment modeling, cross-modal interaction and cyclic permutation from PS6 progressively aggregate multi-modal features.}
    \label{framework}
\end{figure*}

\section{Methodology}\label{sec:Methodology}
As shown in Fig.~\ref{framework}, our framework consists of three key components: a vision feature extractor, Semantic-Driven Token Pruning (SDTP), and Progressive Fusion Network (PFN). Below, we delve into the specific details of each component.

\subsection{Semantic Information Extraction}\label{sec:3_1}
Traditional segmentation methods often fail in person ReID scenarios due to occlusions and background clutter, such as similar-colored clothing or accessories. 
To mitigate this, we employ OpenPifPaf~\cite{kreiss2021openpifpaf} to detect human skeletal keypoints and aggregate the corresponding part-level masks for robust person segmentation. 
For vehicle instances, we utilize SAM2~\cite{ravi2024sam2}, which uses automatically generated prompts proportional to the image dimensions to produce accurate vehicle masks. 
The mask generation is performed offline and remains decoupled from model training and inference, ensuring computational efficiency without incurring additional runtime overhead.

Meanwhile, based on CLIP's vision encoder $\mathcal{F}_{visual}$, we extract features $F_m \in \mathbb{R}^{(N_p+1) \times D}$ from multi-modal images $I_m$ $(m \in \{R, N, T\})$, where $F_m$ contains patch tokens $F^{patch}_m \in \mathbb{R}^{N_p \times D}$ and a global class token $f^{class}_m \in \mathbb{R}^{D}$:
\begin{equation}
F_{m} = [f^{class}_{m}, F^{patch}_{m}] = \mathcal{F}_{visual}(I_{m}).
\end{equation}
Here, $N_p$ represents the number of patch tokens, $D$ is the embedding dimension, and $[\cdot]$ denotes concatenation.
Simultaneously, a semantic mask is generated from $I_{R} \in \mathbb{R}^{3 \times H \times W}$ using $\mathcal{F}_{\text{semantic}}$. This mask is subsequently embedded into patch tokens $M^\prime \in \mathbb{R}^{N_p \times D}$ via a Patch Embedding operation $\mathcal{P}$.
The embedded mask tokens are concatenated with the class token to form the mask-enhanced sequence $M_{m} \in \mathbb{R}^{(N_p+1) \times D}$, formulated as:
\begin{equation}
    M^\prime = \mathcal{P}(\mathcal{F}_{semantic}(I_{R})), \quad
    M_{m} = [f^{class}_{m}, M^\prime].
\end{equation}
Through these operations, we obtain multi-spectral image features and semantic features.

\subsection{Semantic-Driven Token Pruning}
\label{subsec:sdtp}
To leverage rich semantic priors from pre-trained segmentation models, we propose Semantic-Driven Token Pruning (SDTP), a cross-modal mechanism that integrates semantic guidance into multi-spectral feature learning. 
Rather than treating segmentation masks as external dependencies, our framework utilizes them as structured spatial priors to guide feature refinement across all three modalities, enabling robust focus on semantically meaningful regions even under severe appearance degradation.
SDTP adopts a sequential architecture consisting of two stages. 
First, Target-Aware Feature Enhancement (TAFE) establishes fine-grained semantic-spectral alignment through token-level interaction, enhancing discriminative features within foreground regions. 
Subsequently, the Tail Drop Module (TDM) operates on these refined features to suppress low-response tokens in background or noisy areas, effectively pruning irrelevant information and concentrating model capacity on key structural parts.
This cascaded design ensures that token pruning is performed on semantically enriched representations rather than raw feature responses, which are often corrupted under real-world degradations.

\subsubsection{Prompt-S6: Decoupled Prompt Conditioning for Asymmetric State Control}
\label{subsec:ps6}

Building upon the selective state space model (SSM) in Mamba~\cite{gu2023mamba}, we propose Prompt-S6 (PS6), a novel operator designed for efficient cross-modal interaction. Unlike standard Mamba blocks that process a single input modality, PS6 decouples the generation of the SSM's core components to enable conditioning on external auxiliary signals.
Formally, given a primary input token sequence $x \in \mathbb{R}^{N_p \times D}$ (e.g., features from an RGB image), PS6 introduces two auxiliary prompt sequences: $t \in \mathbb{R}^{N_p \times D}$ and $t_p \in \mathbb{R}^{N_p \times D}$, which typically originate from other modalities such as thermal images or semantic masks. The primary sequence $x$ undergoes standard preprocessing via a 1D causal convolution followed by layer normalization, as in the original Mamba architecture.

The key innovation of PS6 lies in the decoupled generation of the SSM projection matrices: the input projection matrix $B$ is derived solely from the gating prompt $t$ through a linear projection, i.e., $B = \mathrm{Linear}_B(t)$; and the output projection matrix $C$ is similarly derived from the projection prompt $t_p$, i.e., $C = \mathrm{Linear}_C(t_p)$. This design allows the external prompts to directly modulate how information from the primary input $x$ is written into and read from the latent state. For instance, a semantic mask prompt can act as a soft gating mechanism, adaptively emphasizing relevant regions and suppressing background noise during state evolution.
This decoupled prompt conditioning mechanism is the core technical contribution of PS6. In this design, external signals independently govern state injection ($B$) and emission ($C$), enabling fine-grained, asymmetric modulation that prior fusion-based or shared-state SSM designs cannot achieve.

This architectural choice is motivated by a fundamental limitation of standard Mamba and its recent multi-modal extensions (e.g., MambaPro~\cite{wang2025mambapro}, S2CrossMamba~\cite{zhang2024s}). These methods integrate multi-modal information either through early feature fusion (e.g., summation or concatenation) or by sharing a single SSM state across modalities. 
In S2CrossMamba, for instance, HSI and LiDAR features are first fused via element-wise addition into a unified representation, which serves as the input to the Cross-SSM. The SSM parameters ($B$, $C$) are then formed as the sum of modality-specific projections, i.e., $B = B_H + B_L$ and $C = C_H + C_L$.
While effective, this approach couples the cross-modal interaction with the primary sequence processing, limiting the ability to apply independent, modality-specific control over how information is written into and read from the latent state. In contrast, PS6 explicitly decouples these roles. The gating prompt $t$ solely governs the writing process via $B = \mathrm{Linear}_B(t)$, while the projection prompt $t_p$ exclusively controls the reading process via $C = \mathrm{Linear}_C(t_p)$. This enables fine-grained, asymmetric modulation of the SSM dynamics by external signals.
By embedding cross-modal interaction directly into the SSM kernel through prompt-decoupled parameterization, PS6 achieves dynamic, context-aware feature modulation with linear computational complexity. This design fundamentally differs from existing multi-modal SSMs that rely on early feature fusion (e.g., concatenation) to jointly generate $B$ and $C$, thereby coupling modality interactions and limiting independent control over state dynamics. The complete procedure is detailed in Algorithm~\ref{alg:ps6}.

\begin{algorithm}[t]
\caption{Prompt-S6 (PS6)}\label{alg:ps6}
\begin{algorithmic}[1]
\Require  
    Primary tokens $x:{\scriptstyle (N_p,D)}$ (main modality), 
    gating prompt $t:{\scriptstyle (N_p,D)}$, projection prompt $t_p:{\scriptstyle (N_p,D)}$
\Ensure $y: (N_p, D)$
\State $x:{\scriptstyle (N_p, D)} \gets \mathrm{Conv1d}(\mathrm{Linear} (x))$ \Comment{Input preprocessing}
\State $B:{\scriptstyle (N_p, K)}$$ \gets \mathrm{Linear_{B}} (t)$ \Comment{From gating prompt}
\State $C:{\scriptstyle (N_p, K)} $$\gets \mathrm{Linear_{C}} (t_p)$ \Comment{From projection prompt}
\State ${/*} \mathrm{Parameter_{\Delta}} \in \mathbb{R}^{(D,D)}{*/}$
\State $\Delta:\scriptstyle (N_p,D)$ $\gets \log(1 + \exp( \mathrm{Linear}_\Delta(x) + \mathrm{Parameter}_{\Delta}))$
\State ${{/*} \mathrm{Parameter_{A}} \in} { \mathbb{R}^{(N_p,1)}} {*/}$
\State $\overline{A},\overline{B} \gets \Delta \otimes \mathrm{Parameter_{A}},\Delta \otimes B$
\State $y: (N_p, D) \gets \mathrm{SSM}(\overline{A}, \overline{B}, C)(x)$
\State \Return $y$
\end{algorithmic}
\end{algorithm}

\subsubsection{Target-Aware Feature Enhancement} 
To enable fine-grained interaction between semantic features and multi-spectral inputs, we propose Target-Aware Feature Enhancement (TAFE). 
This module uses a PS6 module shared across the three modalities to fuse each modality's feature
$F_m$ with its corresponding semantic mask $M_m$, producing an enhanced representation $F^M_m \in \mathbb{R}^{(N_p+1) \times D}$ that incorporates semantic knowledge.
With $\mathrm{LN}$ denoting Layer Normalization~\cite{ba2016layer} for stable interaction, the overall pipeline can be expressed as:
\begin{equation}
F^M_m = \mathrm{PS6}(\mathrm{LN}(F_m), \mathrm{LN}(F_m), M_m).
\end{equation}
The fused feature is then combined with the original input via a skip connection and passed through a Feed-Forward Network (FFN)~\cite{dosovitskiy2020image}, yielding the output $\overline{F}^i_m \in \mathbb{R}^{(N_p+1) \times D}$:
\begin{equation}
\overline{F}^i_m = \mathrm{FFN}(\mathrm{LN}(F_m) + F^M_m),
\end{equation}
where $i$ denotes the $i$-th TAFE layer. TAFE thus enables effective cross-modal interaction and produces semantically enriched representations.

\begin{figure*}
    \centering
    \includegraphics[width=\textwidth]{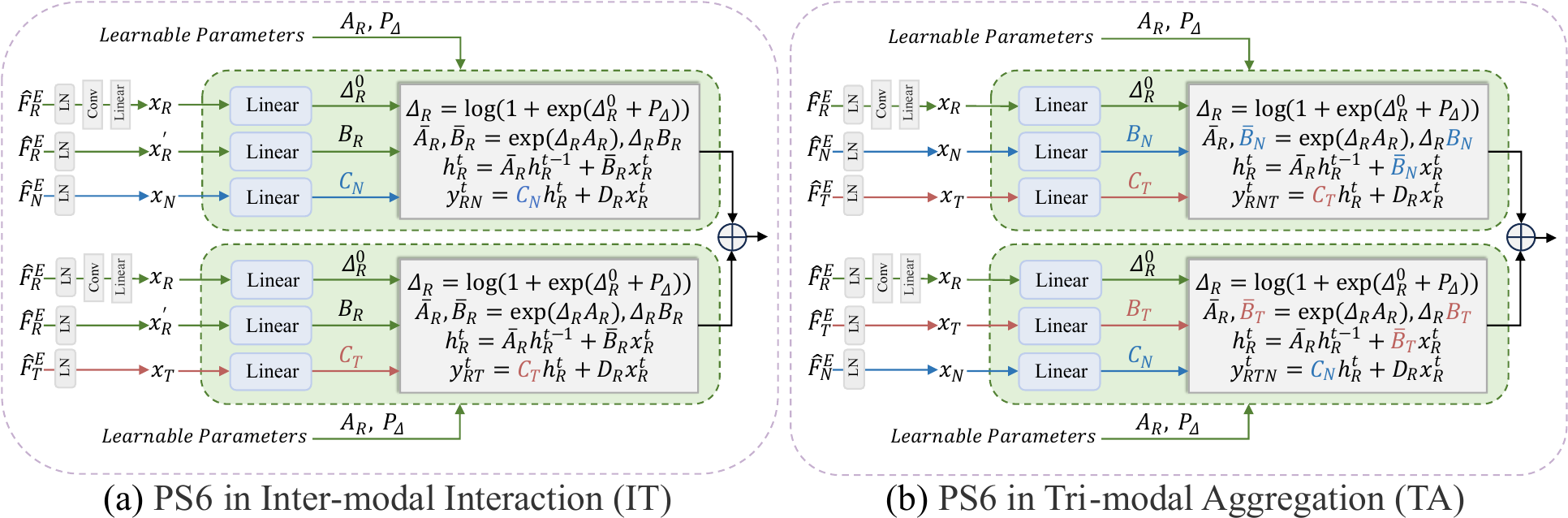}
    \caption{Implementation details of the PFN, illustrated with an RGB modality input $x$. Unlike previous methods that use pairwise interactions to aggregate modalities, TA enables simultaneous interaction among all three modalities, building on initial interactions from Inter-modal Interaction.}
    \label{fig:PFN}
\end{figure*}

\subsubsection{Tail Drop Module} 
This module includes three independent scorers $\Theta_m$ $(m \in \{R, N, T\})$ and a Cross-Modal Union (CMU). After applying a skip connection to retain semantic-guided features:
\begin{equation}
\overline{F}_m = \overline{F}^{i=k}_m + F_m,
\end{equation}
we split $\overline{F}_m$ into $\overline{f}^{class}_m$ and $\overline{F}^{patch}_m$.
Scoring each patch token yields $s^{patch}_m = \Theta_{m}(\overline{F}^{patch}_m)$. 
The top-$k$ scoring tokens in each modality are selected, and the indices $i_R$, $i_N$, and $i_T$ are combined as:
\begin{equation}
    i_m = \mathrm{Topk}(s^{patch}_m), \quad i^{patch} = i_R \cup i_N \cup i_T,
\end{equation}
ensuring cross-modal consistency.
Finally, the critical tokens are retained by applying the fused index to the patch features and merged with $\overline{f}^{class}_m$ to produce $\hat{F}_m \in \mathbb{R}^{(N_p+1) \times D}$:
\begin{equation}
    \hat{F}_m = [\overline{f}^{class}_m,\ (\overline{F}^{patch}_m \odot i^{patch})].
\end{equation}
Unlike methods that retain only a small subset of tokens~\cite{zhang2024magic}, our approach selectively removes the lowest-scoring tokens, preserving a larger share of valuable feature information and maintaining fine-grained details without underutilizing available tokens.
Overall, SDTP leverages soft guidance from non-binary masks through its cascaded architecture, reducing dependence on precise mask boundaries and eliminating the need for part-specific annotations~\cite{zhu2020identity}, which are challenging to obtain under severe imaging degradations.

\subsection{Progressive Fusion Network}
We propose the Progressive Fusion Network (PFN) to model sequential features from multiple modalities. As illustrated in Fig.~\ref{framework}, PFN contains three components: Intra-modal Modeling (IM), Inter-modal Interaction (IT), and Tri-modal Aggregation (TA). IM enhances per-modality features, IT enables cross-modal fusion, and TA integrates all three modalities for comprehensive representation learning.

\subsubsection{Intra-modal Modeling}
Intra-modal Modeling (IM) enhances per-modality features. Given $\hat{F}_m$, the encoder $\mathcal{E}$ outputs refined features $\hat{F}^E_m \in \mathbb{R}^{(N_p+1) \times D}$:
\begin{equation}
    \hat{F}^E_m = \mathcal{E}(\hat{F}_m),
\end{equation}
where $\mathcal{E}$ consists of Multi-Head Self-Attention (MHSA) and a Feed-Forward Network (FFN)~\cite{dosovitskiy2020image}.

\subsubsection{Inter-modal Interaction}
The Inter-modal Interaction (IT) module enables each modality to interact with the other two. For the RGB modality, as shown in Fig.~\ref{fig:PFN}(a), $\hat{F}^E_R$ is paired with $\hat{F}^E_N$ and $\hat{F}^E_T$, respectively, and fed into PS6 for cross-modal interaction:
\begin{equation}
\hat{F}^E_{RN} = \hat{F}^E_R + \mathrm{PS6}(\hat{F}^E_R, \hat{F}^E_R, \hat{F}^E_N),
\end{equation}
\begin{equation}
\hat{F}^E_{RT} = \hat{F}^E_R + \mathrm{PS6}(\hat{F}^E_R, \hat{F}^E_R, \hat{F}^E_T).
\end{equation}
The outputs are averaged to obtain the pairwise fused representation $\hat{F}^{Pairwise}_R \in \mathbb{R}^{(N_p+1) \times D}$:
\begin{equation}
\hat{F}^{Pairwise}_R = (\hat{F}^E_{RN} + \hat{F}^E_{RT}) / 2.
\end{equation}
Here, $\hat{F}^E_{RN}$ and $\hat{F}^E_{RT} \in \mathbb{R}^{(N_p+1) \times D}$ represent the interaction results of RGB with NIR and TIR, respectively. Similarly, we compute the fused features for the NIR and TIR modalities:
\begin{equation}
\hat{F}^{Pairwise}_N = (\hat{F}^E_{NR} + \hat{F}^E_{NT}) / 2,
\end{equation}
\begin{equation}
\hat{F}^{Pairwise}_T = (\hat{F}^E_{TR} + \hat{F}^E_{TN}) / 2.
\end{equation}
For notational brevity, we denote $\hat{F}^{Pairwise}_m$ as $\hat{F}^P_m$ throughout the remainder of the paper.

\subsubsection{Tri-modal Aggregation}
The Tri-modal Aggregation (TA) module uses a shared PS6 block to achieve full tri-modal interaction. Unlike the approach of Wang et al.~\cite{wang2025mambapro}, which concatenates modalities before processing, 
PS6 allows input sequence, SSM input matrix, and output matrix to come from different modalities, as shown in Fig.~\ref{fig:PFN}(b).
Taking RGB modality as an example, we perform two cyclic permutations $(RNT)$ and $(RTN)$ to feed $\hat{F}^P_R$, $\hat{F}^P_N$, and $\hat{F}^P_T$ into PS6. The final representation is obtained via averaging:
\begin{equation}
\hat{F}_{RNT} = \hat{F}^P_R + \mathrm{PS6}(\hat{F}^P_R, \hat{F}^P_N, \hat{F}^P_T),
\end{equation}
\begin{equation}
\hat{F}_{RTN} = \hat{F}^P_R + \mathrm{PS6}(\hat{F}^P_R, \hat{F}^P_T, \hat{F}^P_N),
\end{equation}
\begin{equation}
\hat{F}^{Tri}_R = (\hat{F}_{RNT} + \hat{F}_{RTN}) / 2.
\end{equation}
Similar steps produce $\hat{F}^{Tri}_N$ and $\hat{F}^{Tri}_T$. The class token and averaged patch tokens are concatenated, transformed linearly, and normalized to form the final feature $\hat{f}_m (m \in \{R, N, T\})$:
\begin{equation}
\hat{f}_m = [\hat{f}^{class}_m, \mathcal{L}(\mathcal{A}(\hat{F}^{patch}_m))],
\end{equation}
where $\mathcal{A}(\cdot)$ denotes patch-wise average pooling and $\mathcal{L}(\cdot)$ denotes a linear projection. These are then concatenated to produce the final representation $\hat{f}=[\hat{f}_R, \hat{f}_N, \hat{f}_T] \in \mathbb{R}^{3D}$.
This design enables PFN to capture hierarchical cross-modal interactions while incurring minimal parameter overhead, since PS6 itself is lightweight.

\begin{table}[t]
\caption{Performance comparison on RGBNT201. Methods marked with $\dagger$ are CLIP-based, those with $*$ are ViT-based, while others are CNN-based}
  \label{tab:multi-spectral person ReID}
  \centering
    \renewcommand\arraystretch{1.3}
 \setlength{\tabcolsep}{9.15pt}
  \begin{tabular}{cr|cccc}
      \noalign{\hrule height 1pt}
  &\multicolumn{1}{c|}{\multirow{2}{*}{\textbf{Methods}}}   & \multicolumn{4}{c}{\textbf{RGBNT201}} \\ \cline{3-6}
  & & \textbf{mAP} & \textbf{R-1} & \textbf{R-5} & \textbf{R-10} \\ \hline
  \multirow{3}{*}{\rotatebox{90}{\textbf{Single}}}
   &OSNet~\cite{zhou2019omni}  & 25.4 & 22.3 & 35.1 & 44.7 \\
   &CAL~\cite{rao2021counterfactual}  & 27.6 & 24.3 & 36.5 & 45.7 \\
   &PCB~\cite{sun2018beyond}  & 32.8 & 28.1 & 37.4 & 46.9 \\ \hline
  \multirow{15}{*}{\rotatebox{90}{\textbf{Multi-Modal}}}
  & HAMNet~\cite{li2020multi}   & 27.7         & 26.3            & 41.5            & 51.7             \\
  & PFNet~\cite{zheng2021robust}    & 38.5         & 38.9            & 52.0            & 58.4             \\
  & IEEE~\cite{wang2022interact}     & 47.5         & 44.4            & 57.1            & 63.6             \\
  & DENet~\cite{zheng2023dynamic}    & 42.4         & 42.2            & 55.3            & 64.5            \\
  & LRMM~\cite{wu2025lrmm} & 52.3 & 53.4 & 64.6 & 73.2\\
  & UniCat$^*$~\cite{crawford2023unicat}   & 57.0         & 55.7            & -            & -            \\
& HTT$^*$~\cite{wang2024heterogeneous} &71.1 &73.4 &83.1 &87.3\\
& TOP-ReID$^*$~\cite{wang2024top}  &72.3 &76.6 &84.7 &89.4\\
& EDITOR$^*$~\cite{zhang2024magic} & 66.5       & 68.3           & 81.1        & 88.2             \\
& RSCNet$^*$~\cite{yu2024representation} & 68.2 & 72.5 & - & - \\
& WTSF-ReID$^*$~\cite{yu2025wtsf} & 67.9 &72.2 &83.4 &89.7 \\
& DESANet$^*$~\cite{dong2025escaping} & 74.6 &77.6 &87.1 &91.3 \\
& PromptMA$^\dagger$~\cite{zhang2025prompt} & 78.4 &80.9 &87.0 &88.9 \\
& MambaPro$^\dagger$~\cite{wang2025mambapro} & 78.9 & 83.4 & {89.8} & 91.9 \\
& DeMo$^\dagger$~\cite{wang2025decoupled}  &{79.0} 	 &{82.3} 	 &88.8 	 &{92.0}      \\
& IDEA w/o Text$^\dagger$~\cite{wang2025idea}  &74.5 	 &75.0	 &84.8 	 &88.8     \\
& IDEA$^\dagger$~\cite{wang2025idea}  &{80.2} 	 &82.1 	 &{90.0} 	 &{93.3}      \\
\rowcolor[gray]{0.92}
  & $\mathrm{\textbf{PRISM}}^\dagger$  &80.5 	 &84.0 	 &{91.7} 	 &93.9      \\
  \noalign{\hrule height 1pt}
  \end{tabular}
\end{table}

\subsection{Objective Functions}
As depicted in Fig.~\ref{framework}, the objective function of our model consists of two components: losses associated with the image encoder and the Progressive Fusion Network (PFN). Both the backbone network and the PFN are supervised using a combination of label smoothing cross-entropy loss~\cite{szegedy2016rethinking} and triplet loss~\cite{hermans2017defense}. The global loss for each component can be expressed as:
\begin{equation}
\mathcal{L}_g(\mathcal{X}) = \mathcal{L}_{ce}(\mathcal{X}) + \mathcal{L}_{tri}(\mathcal{X}),
\end{equation}
where $ \mathcal{X} $ denotes input features. Finally, the overall loss of our framework can be expressed as:
\begin{equation}
\mathcal{L} = \mathcal{L}_g([f^{class}_{R},f^{class}_{N},f^{class}_{T}]) + \mathcal{L}_g(\hat{f}).
\end{equation}

\section{Experiment}\label{sec:Ex}
\begin{table}[t]
\caption{Performance comparison on RGBNT100 and MSVR310}
    \label{tab:multi-spectral vehicle ReID}
    \centering
    \renewcommand\arraystretch{1.3}
    \setlength{\tabcolsep}{9.75pt}
    \begin{tabular}{cr|cccc}
        \noalign{\hrule height 1pt}
    &\multicolumn{1}{c|}{\multirow{2}{*}{\textbf{Methods}}} &  \multicolumn{2}{c}{\textbf{RGBNT100}} & \multicolumn{2}{c}{\textbf{MSVR310}} \\\cline{3-6}
    & & \textbf{mAP} & \textbf{R-1} & \textbf{mAP} & \textbf{R-1} \\
    \hline
    \multirow{4}{*}{\rotatebox{90}{\textbf{Single}}}
    &PCB~\cite{sun2018beyond}& 57.2 & 83.5 & 23.2 & 42.9 \\
    &OSNet~\cite{zhou2019omni}& 75.0 & 95.6 & 28.7 & 44.8 \\
    &AGW~\cite{ye2021deep} & 73.1 & 92.7 & 28.9 & 46.9 \\
    &TransReID$^*$~\cite{he2021transreid}& 75.6 & 92.9 & 18.4 & 29.6 \\
    \hline
    \multirow{18}{*}{\rotatebox{90}{\textbf{Multi-Modal}}}
    &HAMNet~\cite{li2020multi} & 74.5 & 93.3 & 27.1 & 42.3 \\
    &PFNet~\cite{zheng2021robust}& 68.1 & 94.1 & 23.5 & 37.4 \\
    &GAFNet~\cite{guo2022generative} & 74.4 & 93.4 & - & - \\
    &GPFNet~\cite{he2023graph} & 75.0 & 94.5 & - & - \\
    &CCNet~\cite{zheng2023cross} & 77.2 & 96.3 & 36.4 & 55.2 \\
    & LRMM~\cite{wu2025lrmm} & 78.6 & 96.7 & 36.7 &49.7\\
    &GraFT$^*$~\cite{yin2023graft} &76.6 &94.3 &- &-\\
    &UniCat$^*$~\cite{crawford2023unicat}    & 79.4         & 96.2  & -            & -            \\
    &PHT$^*$~\cite{pan2023progressively} & 79.9 & 92.7 & - & - \\
    & HTT$^*$~\cite{wang2024heterogeneous} &75.7&92.6&- &-\\
    & TOP-ReID$^*$~\cite{wang2024top} &81.2 & 96.4 & 35.9 & 44.6 \\
    & EDITOR$^*$~\cite{zhang2024magic} & 82.1 & 96.4 &39.0 & 49.3\\
    & FACENet$^*$~\cite{zheng2025flare} & 81.5 &96.9 &36.2 &54.1 \\
    & RSCNet$^*$~\cite{yu2024representation} &82.3 &96.6 &39.5 &49.6\\
    & WTSF-ReID$^*$~\cite{yu2025wtsf} & 82.2 &96.5 & 39.2 & 49.1 \\
    & DESANet$^*$~\cite{dong2025escaping} & 82.1 & {97.4} &39.2 &47.8 \\
    & PromptMA$^\dagger$~\cite{zhang2025prompt} & 85.3 & {97.4} &{55.2} &{64.5} \\
    & MambaPro$^\dagger$~\cite{wang2025mambapro} & 83.9 & 94.7 &47.0 & 56.5 \\
    & DeMo$^\dagger$~\cite{wang2025decoupled} &{86.2} 	&  {97.6} &{49.2}	&59.8 \\
    & IDEA$^\dagger$~\cite{wang2025idea}&   {87.2} 	&96.5 &47.0	&62.4 \\
    \rowcolor[gray]{0.92}& $\mathrm{\textbf{PRISM}}^\dagger$&86.1 	&97.8 &47.6	&{64.8} \\
    \noalign{\hrule height 1pt}
    \end{tabular}
\end{table}

\begin{table}[t]
    \centering
    \caption{Performance comparison on WMVEID863}
    \label{tab:wmveid863_results}
    \renewcommand\arraystretch{1.3}
    \setlength{\tabcolsep}{9.75pt}
    \begin{tabular}{cr|cccc}
        \noalign{\hrule height 1pt}
        &\multicolumn{1}{c|}{\multirow{2}{*}{\textbf{Methods}}} & \multicolumn{4}{c}{\textbf{WMVEID863}} \\ \cline{3-6}
        & & \textbf{mAP} & \textbf{R-1} & \textbf{R-5} & \textbf{R-10} \\ \hline
        \multirow{9}{*}{\rotatebox{90}{\textbf{Multi-Modal}}}
        & HAMNet~\cite{li2020multi}           & 45.6 & 48.5 & 63.1 & 68.8 \\
        & PFNet~\cite{zheng2021robust}        & 50.1 & 55.9 & 68.7 & 75.1 \\
        & IEEE~\cite{wang2022interact}                & 45.9 & 48.6 & 64.3 & 67.9 \\
        & CCNet~\cite{zheng2023cross}         & 50.3 & 52.7 & 69.6 & 75.1 \\ 
        & EDITOR$^*$~\cite{zhang2024magic}       & 65.6 & 73.8 & 80.0 & 82.3 \\
        & TOP-ReID$^*$~\cite{wang2024top}     & 67.7 & 75.3 & 80.8 & 83.5 \\
        & FACENet$^*$~\cite{zheng2025flare}     & 69.8 & 77.0 & 81.0 & 84.2 \\
        & MambaPro$^\dagger$~\cite{wang2025mambapro} & 69.8 & 76.3 & 81.3 & 86.1 \\
        & DeMo$^\dagger$~\cite{wang2025decoupled}   & 68.9 & 75.8 & 81.1 & 84.3 \\
        \rowcolor[gray]{0.92}
        & $\mathrm{\textbf{PRISM}}^\dagger$ & 70.7 & 78.1 & 84.5 & 87.9 \\
        \noalign{\hrule height 1pt}
    \end{tabular}
\end{table}

\begin{table}[t]
  \caption{Performance comparison of PS6 against recent SSM variants}
  \label{tab:ssm}
  \centering
  \renewcommand\arraystretch{1.18}
  \setlength{\tabcolsep}{14.65pt}
  \begin{tabular}{r|ccc}
      \noalign{\hrule height 1pt}
      \textbf{Modules} & GFLOPs & Params & mAP \\\hline
      PS6 (Ours) & 0.337 & 0.79 & 80.5 \\
      Mamba~\cite{gu2023mamba} & 0.472 & 2.17 & 79.3 \\
      VMamba~\cite{liu2024vmamba} & 0.511 &2.17  & 79.0 \\
      Vision Mamba~\cite{zhu2024vision} & 0.944 &2.17  & 78.5 \\
      CrossMamba~\cite{he2025pan} & 0.473 &2.18  & 77.7 \\
      \noalign{\hrule height 1pt}
  \end{tabular}
\end{table}

\subsection{Experimental Setup}\label{sec:4_1}
\subsubsection{Datasets and Evaluation Protocols}
To comprehensively evaluate our method, we conduct experiments on four multi-modal object ReID benchmarks. 
For generating high-quality masks, we employ semantic keypoint detection~\cite{kreiss2021openpifpaf} and SAM2~\cite{ravi2024sam2} to obtain masks for persons and vehicles, respectively. 
To be specific, RGBNT201~\cite{zheng2021robust} is a person ReID dataset comprising RGB, NIR, and TIR images, featuring 4,787 triplets with 14,361 annotations. 
RGBNT100~\cite{li2020multi} is a large-scale vehicle ReID dataset with 17,250 triplets and 51,750 annotations, incorporating challenges such as occlusion and abnormal lighting. 
MSVR310~\cite{zheng2023cross} is a smaller yet more complex vehicle ReID dataset, containing 2,087 triplets and 6,261 annotations with intricate visual challenges.
WMVEID863~\cite{zheng2025flare} is a vehicle ReID dataset containing 4,709 triplets and 14,127 annotations, specifically emphasizing motion blur and strong glare under real-world surveillance conditions.
For evaluation metrics, performance is measured using mean Average Precision (mAP) and Cumulative Matching Characteristic (CMC) at Rank-\textit{K} (\textit{K} = 1, 5, 10), reported in percentage (\%). 
Specifically, final results are presented as \textit{mAP (\%)}, \textit{Rank-1 (\%)}, \textit{Rank-5 (\%)}, and \textit{Rank-10 (\%)}.

\subsubsection{Implementation Details}
Our framework is implemented using PyTorch on an NVIDIA A6000 GPU, leveraging CLIP~\cite{radford2021learning} as the visual backbone. 
For RGBNT201, input images are resized to $256 \times 128$, while for RGBNT100 and MSVR310, they are resized to $128 \times 256$. 
Data augmentation techniques include random horizontal flipping, cropping, and erasing \cite{zhong2020random}, which are employed to enhance model robustness.
We adopt the same loss function as DeMo~\cite{wang2025decoupled} to train the feature extractor and subsequent modules, where both components are supervised using a combination of label smoothing cross-entropy loss~\cite{szegedy2016rethinking} and triplet loss~\cite{hermans2017defense}.
During training, we adopt a mini-batch size of 64 for RGBNT201 and MSVR310, sampling 8 images per identity, and a mini-batch size of 128 for RGBNT100, sampling 16 images per identity. 
The total number of training epochs is set to 60 for RGBNT201 and MSVR310 and 50 for RGBNT100. 
The model is optimized using the Adam optimizer with a learning rate of $3.5e^{-4}$, while the visual encoder uses a smaller learning rate of $5e^{-6}$.

\begin{table}[t]
  \caption{Performance comparison (mAP) under different pixel-level modality misalignment. The column headers indicate the degree of random spatial shift (in pixels) applied to the auxiliary modality}
  \label{tab:pixel_offset}
  \centering
  \renewcommand\arraystretch{1.18}
  \setlength{\tabcolsep}{7.5pt}
  \begin{tabular}{r|cccccc}
      \noalign{\hrule height 1pt}
      \textbf{Methods} & - & 10 & 20 & 30 & 40 & 50 \\\hline
      PromptMA$^\dagger$~\cite{zhang2025prompt} & 78.4 & 78.1 &77.2 &74.2 &70.8 &66.1 \\
      DeMo$^\dagger$~\cite{wang2025decoupled} &79.0 &79.4 &78.2 &75.1 &71.3 &65.4 \\ 
      IDEA$^\dagger$~\cite{wang2025idea} &80.2 &81.3 &79.5 &76.7 &71.4 &66.1 \\
      MFRNet$^\dagger$~\cite{feng2025multi} & 80.7 &79.7 &78.1 &74.8 &71.1 &66.4 \\
    \rowcolor[gray]{0.92} $\mathrm{\textbf{PRISM}}^\dagger$&80.5 &81.9 &80.1 &77.0 &71.8 &66.6 \\
      \noalign{\hrule height 1pt}
  \end{tabular}
\end{table}

\begin{table}[t]
  \caption{Performance comparison (mAP) under different levels of image occlusion. The column headers indicate the ratio of the occluded area relative to the total image area}
  \label{tab:occluded_area}
  \centering
  \renewcommand\arraystretch{1.18}
  \setlength{\tabcolsep}{10.0pt}
  \begin{tabular}{r|ccccc}
      \noalign{\hrule height 1pt}
      \textbf{Methods} & - & 0.1 & 0.2 & 0.3 & 0.4 \\\hline
      PromptMA$^\dagger$~\cite{zhang2025prompt} & 78.4 & 74.3 &64.8 &58.4 &45.5 \\
      DeMo$^\dagger$~\cite{wang2025decoupled} &79.0 &73.5 &65.1 &61.4 &48.0 \\ 
      IDEA$^\dagger$~\cite{wang2025idea} &80.2 &76.1 &68.2 &63.9 &49.8 \\
      MFRNet$^\dagger$~\cite{feng2025multi} & 80.7 &75.9 &67.7 &62.1 &49.3 \\
    \rowcolor[gray]{0.92} $\mathrm{\textbf{PRISM}}^\dagger$&80.5 &77.0 &68.5 &62.4 &49.7 \\
      \noalign{\hrule height 1pt}
  \end{tabular}
\end{table}

\begin{table}[t]
  \caption{Performance comparison (R-1) under different pixel-level modality misalignment. The column headers indicate the degree of random spatial shift (in pixels) applied to the auxiliary modality}
  \label{tab:pixel_offset_r1}
  \centering
  \renewcommand\arraystretch{1.18}
  \setlength{\tabcolsep}{7.5pt}
  \begin{tabular}{r|cccccc}
      \noalign{\hrule height 1pt}
      \textbf{Methods} & - & 10 & 20 & 30 & 40 & 50 \\\hline
      PromptMA$^\dagger$~\cite{zhang2025prompt} & 80.9 & 81.0 & 80.6 & 80.4 & 78.6 & 78.0 \\
      DeMo$^\dagger$~\cite{wang2025decoupled} & 82.3 & 81.8 & 81.0 & 80.3 & 78.8 & 77.8 \\ 
      IDEA$^\dagger$~\cite{wang2025idea} & 82.1 & 84.8 & 84.6 & 81.7 & 79.2 & 78.1 \\
      MFRNet$^\dagger$~\cite{feng2025multi} & 83.6 & 83.3 & 83.0 & 81.2 & 81.8 & 79.1 \\
    \rowcolor[gray]{0.92} $\mathrm{\textbf{PRISM}}^\dagger$ & 84.0 & 85.4 & 84.6 & 84.7 & 83.6 & 82.1 \\
      \noalign{\hrule height 1pt}
  \end{tabular}
\end{table}

\begin{table}[t]
  \caption{Performance comparison (R-1) under different levels of image occlusion. The column headers indicate the ratio of the occluded area relative to the total image area}
  \label{tab:occluded_area_r1}
  \centering
  \renewcommand\arraystretch{1.18}
  \setlength{\tabcolsep}{10.0pt}
  \begin{tabular}{r|ccccc}
      \noalign{\hrule height 1pt}
      \textbf{Methods} & - & 0.1 & 0.2 & 0.3 & 0.4 \\\hline
      PromptMA$^\dagger$~\cite{zhang2025prompt} & 80.9 & 79.1 & 70.7 & 67.7 & 61.2 \\
      DeMo$^\dagger$~\cite{wang2025decoupled} & 82.3 & 77.6 & 70.3 & 68.1 & 65.0 \\ 
      IDEA$^\dagger$~\cite{wang2025idea} & 82.1 & 80.9 & 74.4 & 72.8 & 65.7 \\
      MFRNet$^\dagger$~\cite{feng2025multi} & 83.6 & 81.2 & 74.9 & 72.7 & 66.6 \\
    \rowcolor[gray]{0.92} $\mathrm{\textbf{PRISM}}^\dagger$ & 84.0 & 83.0 & 78.1 & 74.4 & 68.1 \\
      \noalign{\hrule height 1pt}
  \end{tabular}
\end{table}

\begin{figure}[t]
  \centering
    {
  \includegraphics[width=\linewidth]{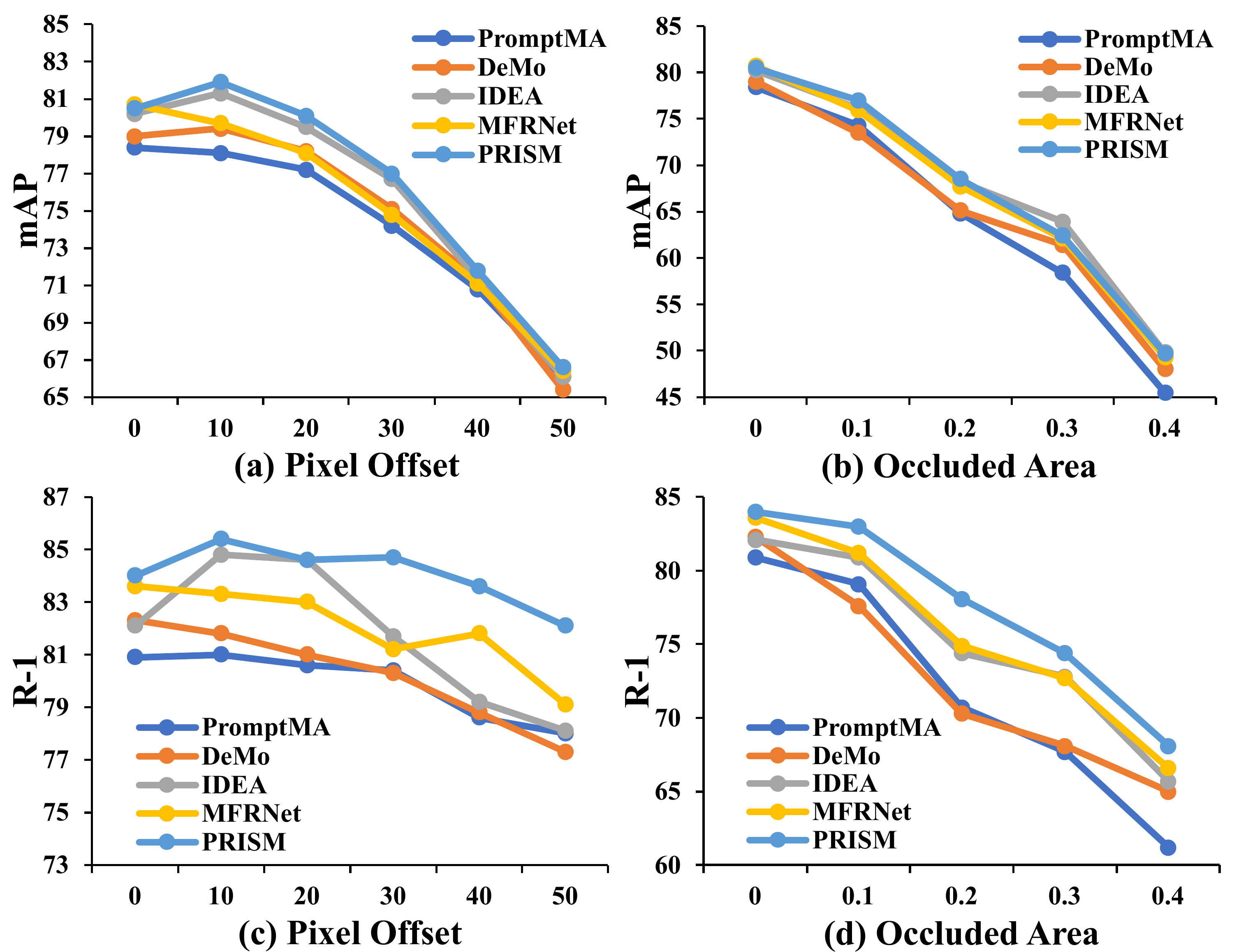}
  }
  \vspace{-3mm}
   \caption{Performance evaluation under challenging conditions on the RGBNT201. 
    (a) Robustness to pixel-level modality misalignment (spatial shift in pixels, mAP). 
    (b) mAP degradation under varying levels of image occlusion (relative occluded area ratio). 
    (c) Robustness to pixel-level modality misalignment (R-1). 
    (d) R-1 degradation under image occlusion.}
  \label{fig:robustness}
\end{figure}

\begin{table}[t]
  \caption{Performance comparison with different modules in PRISM}
  \label{tab:main_ablation}
  \centering
  \renewcommand\arraystretch{1.0}
  \setlength{\tabcolsep}{10.75pt}
  \begin{tabular}{cccccc}
      \noalign{\hrule height 1pt}
      \multicolumn{1}{c}{\multirow{2}{*}{\textbf{Index}}} &\multicolumn{3}{c}{\textbf{Modules}} & \multicolumn{2}{c}{\textbf{Metrics}} \\
      \cmidrule(r){2-4} \cmidrule(r){5-6}
 & \textbf{Mask}              & \textbf{SDTP}                & \textbf{PFN}                   & \textbf{mAP}    & \textbf{Rank-1}   \\\hline
  A                  & -                  & \XSolidBrush                  & \XSolidBrush                    & 70.5  & 73.4 \\
  B                  & -                  & \XSolidBrush                  & \CheckmarkBold                      & 73.7  & 78.1 \\
  \multirow{1}{*}{C} & \multirow{1}{*}{\XSolidBrush} & \multirow{1}{*}{\CheckmarkBold} & \multirow{1}{*}{\CheckmarkBold}    & 75.8  & 79.3 \\
  \rowcolor[gray]{0.92}
  \multirow{1}{*}{D} & \multirow{1}{*}{\CheckmarkBold} & \multirow{1}{*}{\CheckmarkBold} & \multirow{1}{*}{\CheckmarkBold}    &\textbf{80.5} &\textbf{84.0}  \\
  \noalign{\hrule height 1pt}
  \end{tabular}
\end{table}

\begin{table}[t]
  \caption{Comparison with different components in SDTP}
  \label{tab:SDTP_ablation}
  \centering
  \renewcommand\arraystretch{1.03}
  \setlength{\tabcolsep}{8.5pt}
  \begin{tabular}{cccccc}
      \noalign{\hrule height 1pt}
      \multicolumn{1}{c}{\multirow{2}{*}{\textbf{Index}}} &\multicolumn{3}{c}{\textbf{SDTP}} & \multicolumn{2}{c}{\textbf{Metrics}} \\
      \cmidrule(r){2-4} \cmidrule(r){5-6}
& \textbf{TAFE}              & \textbf{TDM}                & \textbf{PS6 Quantity}                   & \textbf{mAP}    & \textbf{Rank-1}   \\\hline
A                   & \XSolidBrush                  & \XSolidBrush                  & - & 73.7  & 78.1   \\
B                  & \CheckmarkBold                  & \XSolidBrush                 & 2                     & 76.4  & 80.7                    \\
C                  &\CheckmarkBold                 & \CheckmarkBold                  & 1                      & 77.5  & 79.1                \\
D                  & \CheckmarkBold                & \CheckmarkBold                & 3                      & 79.3  & 82.2                \\
\rowcolor[gray]{0.92}
E & \multirow{1}{*}{\CheckmarkBold} & \multirow{1}{*}{\CheckmarkBold} & \multirow{1}{*}{2}     &\textbf{80.5}  &\textbf{84.0}   \\
  \noalign{\hrule height 1pt}
  \end{tabular}
\end{table}

\begin{table}[t]
  \caption{Comparison with different components in PFN}
  \label{tab:PFN_ablation}
  \centering
  \renewcommand\arraystretch{1.0}
  \setlength{\tabcolsep}{12pt}
  \begin{tabular}{cccccc}
      \noalign{\hrule height 1pt}
      \multicolumn{1}{c}{\multirow{2}{*}{\textbf{Index}}} &\multicolumn{3}{c}{\textbf{PFN}} & \multicolumn{2}{c}{\textbf{Metrics}} \\
      \cmidrule(r){2-4} \cmidrule(r){5-6}
& \textbf{Intra}              & \textbf{Inter}                & \textbf{Tri}                   & \textbf{mAP}    & \textbf{Rank-1}   \\\hline
A                   & \XSolidBrush                  & \XSolidBrush                 & \XSolidBrush & 76.3  & 79.9   \\
B                  & \CheckmarkBold                 & \XSolidBrush                  & \XSolidBrush                    & 76.4  & 81.3                    \\
C                  & \CheckmarkBold                 & \CheckmarkBold                 & \XSolidBrush                     & 77.6  & 79.4                \\
\rowcolor[gray]{0.92}
D & \multirow{1}{*}{\CheckmarkBold} & \multirow{1}{*}{\CheckmarkBold} & \multirow{1}{*}{\CheckmarkBold}     &\textbf{80.5}  &\textbf{84.0}   \\
  \noalign{\hrule height 1pt}
  \end{tabular}
\end{table}

\begin{table}[t]
  \caption{Comparison of model performance using different masks generated on RGBNT201. RGB, NIR, and TIR denote using a mask generated from that single modality for all inputs; R/N/T denotes using modality-specific masks}
  \label{tab:data}
  \centering
  \renewcommand\arraystretch{1.18}
  \setlength{\tabcolsep}{10.25pt}
  \begin{tabular}{ccccccc}
      \noalign{\hrule height 1pt}
      \textbf{Data} & \textbf{Metrics} & \textbf{RGB} & \textbf{NIR} & \textbf{TIR} & \textbf{R/N/T} \\\hline
      \textbf{OpenPifPaf} & mAP & 80.5 & 76.9 & 79.6 & 80.1 \\
      \cite{kreiss2021openpifpaf} & Rank-1 & 84.0 & 81.5 & 83.4 & 84.1 \\\noalign{\hrule height 1pt}
      \textbf{Data}&\textbf{Metrics}&\multicolumn{2}{c}{\textbf{Heatmap}} & \multicolumn{2}{c}{\textbf{Binary Mask}}\\\hline
      \textbf{SAM2} & mAP &\multicolumn{2}{c}{79.4} & \multicolumn{2}{c}{78.2} \\
      ~\cite{ravi2024sam2} & Rank-1 & \multicolumn{2}{c}{85.4} & \multicolumn{2}{c}{83.1} \\
      \noalign{\hrule height 1pt}
  \end{tabular}
\end{table}

\begin{table}[t]
  \caption{Comparison of model performance using different masks generated on RGBNT100}
  \label{tab:mask4100}
  \centering
  \renewcommand\arraystretch{1.18}
 \setlength{\tabcolsep}{13.0pt}
  \begin{tabular}{r|cccc}
      \noalign{\hrule height 1pt}
      \textbf{Data} & \textbf{mAP} & \textbf{R-1} & \textbf{R-5} & \textbf{R-10} \\
      \hline
      \textbf{OpenPifPaf}~\cite{kreiss2021openpifpaf} & 82.9 & 93.4 & 94.5 & 95.6 \\
      \textbf{SAM2}~\cite{ravi2024sam2} & 86.1 &97.8  & 98.2 & 98.6 \\
      \noalign{\hrule height 1pt}
  \end{tabular}
\end{table}

\begin{table}[t]
  \caption{Performance and efficiency comparison of different mask generation strategies on RGBNT201}
  \label{tab:inf_supp}
  \centering
  \renewcommand\arraystretch{1.18}
  \setlength{\tabcolsep}{2.65pt}
  \begin{tabular}{r|ccc}
      \noalign{\hrule height 1pt}
      \textbf{Mask generator G} & Inf time of G (s) & mAP & Rank-1 \\\hline
      Ours w/o SDTP & - & 73.7 & 78.1 \\
      Ours w/o Mask & - & 75.8 & 79.3 \\
      SAM w/ Box Prompt & 0.0238 &75.6  & 78.6 \\
      SAM2 Binary Mask w/o Box Prompt & 0.0113 &76.4  & 79.1 \\
      SAM w/ SamAutomaticMaskGenerator & 0.1097 &77.5  & 82.3 \\
      SAM2 w/ Dynamic Box Prompt (Ours) & 0.0115 &80.5  & 85.9 \\
      \noalign{\hrule height 1pt}
  \end{tabular}
\end{table}

\begin{table}[t]
  \caption{Performance comparison (mAP) of unified versus separate token selection under varying retention rates. 
The unified strategy achieves the highest mAP at a retention rate of 0.5}
  \label{tab:thyper}
  \centering
  \renewcommand\arraystretch{1.18}
 \setlength{\tabcolsep}{10.25pt}
  \begin{tabular}{r|ccccc}
      \noalign{\hrule height 1pt}
      \textbf{Retention Rate} & \textbf{0.3} & \textbf{0.4} & \textbf{0.5} & \textbf{0.6} & \textbf{0.7} \\\hline
      \textbf{Union } & 78.1 & 79.6 & 80.5 & 80.0 & 79.1 \\
      \textbf{Separate } & 77.9 &79.4  & 79.7 & 78.7 & 78.6 \\\noalign{\hrule height 1pt}
  \end{tabular}
\end{table}

\begin{figure*}[t]
  \centering
  \includegraphics[width=1.0\linewidth]{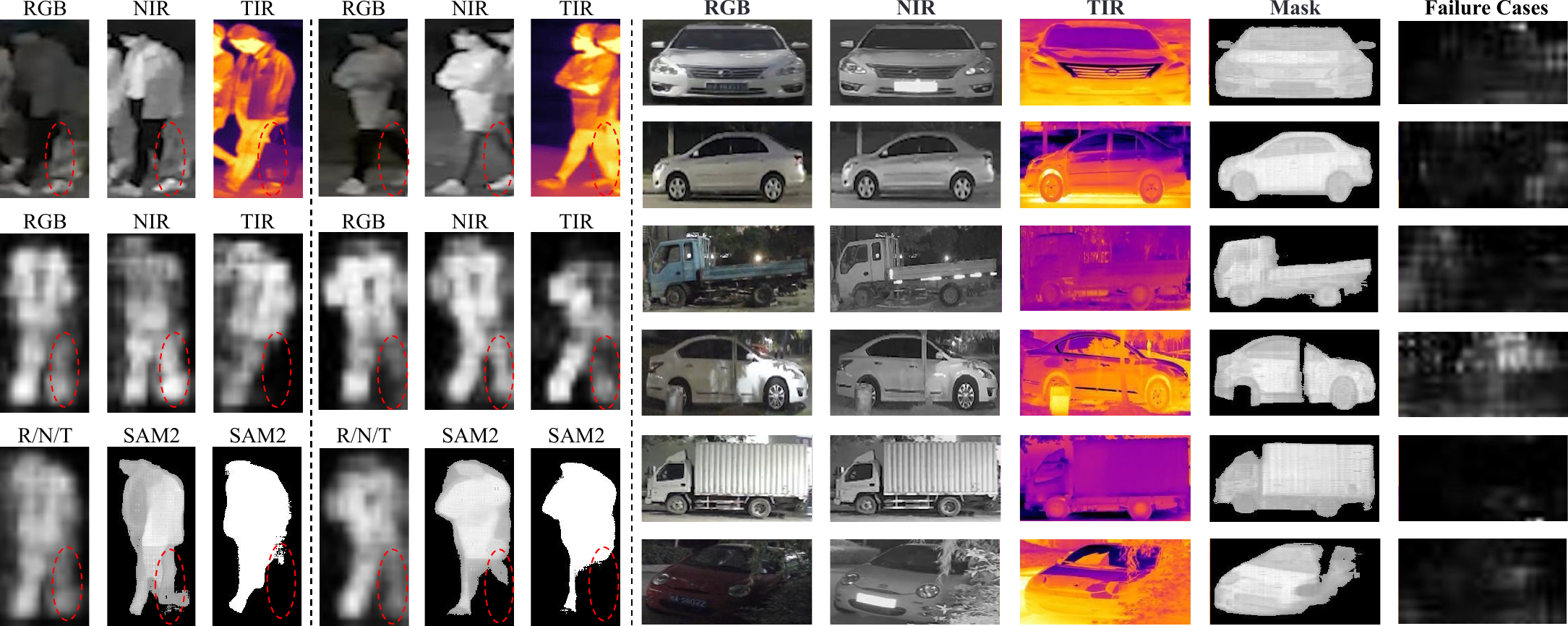}
   \caption{Qualitative comparison of semantic masks generated on the RGBNT201 and RGBNT100 datasets. 
Our mask generation approach produces more accurate and robust masks under imaging degradations such as low lighting, occlusion, and thermal variations.}
  \label{fig:mask}
\end{figure*}

\subsection{Comparison with State-of-the-Art Methods}\label{sec:sota}
\subsubsection{Multi-Modal Person ReID}
In Table~\ref{tab:multi-spectral person ReID}, we compare PRISM$^\dagger$ with existing multi-modal approaches on the RGBNT201 dataset. Our model, enhanced with PS6, surpasses existing state-of-the-art (SOTA) methods across all metrics. 
Specifically, PRISM$^\dagger$ obtains 80.5\% mAP and 84.0\% Rank-1 accuracy, outperforming TOP-ReID$^*$~\cite{wang2024top} by 8.2\% in mAP and 7.4\% in Rank-1.
Compared to other CLIP-based SOTA methods such as DeMo$^\dagger$~\cite{wang2025decoupled} and IDEA$^\dagger$~\cite{wang2025idea}, our method consistently achieves superior results across all evaluation metrics. 
These findings highlight the effectiveness of integrating semantic masks for enhancing feature discrimination.
Furthermore, our method achieves 80.5\% mAP with 109.25M trainable parameters, demonstrating strong performance under a compact model footprint. 
It surpasses TOP-ReID~\cite{wang2024top} (72.3\% mAP, 324.53M parameters), EDITOR~\cite{zhang2024magic} (66.5\% mAP, 118.55M), WTSF-ReID~\cite{yu2025wtsf} (67.9\% mAP, 143.60M), and RSCNet~\cite{yu2024representation} (68.2\% mAP, 124.10M) in accuracy, despite using significantly fewer parameters. 
This highlights the superior parameter efficiency of the proposed approach.

\subsubsection{Multi-Modal Vehicle ReID}
We evaluate PRISM$^\dagger$ against SOTA methods on the RGBNT100 and MSVR310 datasets, as shown in Table~\ref{tab:multi-spectral vehicle ReID}. 
On RGBNT100, PRISM$^\dagger$ achieves an mAP of {86.1\%}, improving upon EDITOR$^*$~\cite{zhang2024magic} by {4.0\%} in mAP. 
On the more challenging MSVR310 dataset, our model attains 47.6\% mAP and 64.8\% Rank-1 accuracy, surpassing EDITOR$^*$ by 8.6\% in mAP and 15.5\% in Rank-1.
Furthermore, as shown in Table~\ref{tab:wmveid863_results}, PRISM attains 70.7\% mAP and 78.1\% Rank-1 on the WMVEID863 dataset, surpassing FACENet$^*$~\cite{zheng2025flare} by 0.9\% and 1.1\%, respectively, and establishing SOTA performance across all evaluation metrics.
This highlights the effectiveness of PRISM's fine-grained token pruning in handling localized disturbances such as motion blur and intense flare.
These results demonstrate that PRISM achieves competitive performance and exhibits strong generalization capability across diverse and challenging multi-spectral vehicle ReID scenarios.

\subsubsection{Comparison with Recent SSM Variants}
PS6 is grounded in State Space Models (SSMs) theory, and its core contribution lies in a simple yet effective architecture for unified multi-modal feature processing, distinguishing it from existing Mamba variants. 
We compare PS6 against Inter-Mamba~\cite{wang2025mambapro}, attention-based methods~\cite{zhang2024magic}, and Mamba baselines (Mamba, Mamba*), demonstrating superior computational efficiency, as shown in Fig.~\ref{fig1}(b). 
Specifically, Mamba and Mamba* denote feature aggregation via summing and concatenating, respectively, after Mamba-based self-interactions.
Furthermore, comparisons with other representative Mamba variants on the RGBNT201 dataset highlight the overall performance advantage of PS6, as shown in Table~\ref{tab:ssm}.

\subsubsection{Robustness Evaluation}
To comprehensively evaluate the robustness of the proposed PRISM framework, we conduct dedicated experiments under two challenging scenarios: partial occlusion and inter-modality misalignment, using the RGBNT201 dataset. 
For partial occlusion, we simulate realistic blocking by applying random square masks of varying sizes to the same spatial location across all three modalities (RGB, NIR, and TIR). 
To model modality misalignment, we fix the TIR image as the reference and independently apply random horizontal and vertical shifts to the RGB and NIR images, thereby emulating registration errors commonly encountered in practical multi-spectral capture systems. 
As shown in Fig.~\ref{fig:robustness} and summarized in Tables~\ref{tab:pixel_offset}-\ref{tab:occluded_area_r1}, PRISM demonstrates consistently stable performance under both degradation types.
Notably, while MFRNet~\cite{feng2025multi} slightly outperforms PRISM in mAP under clean conditions (80.7 versus 80.5), our method exhibits superior robustness when data quality is compromised. 
An interesting observation from the misalignment study is that a small offset (e.g., 10 pixels) often leads to a slight mAP improvement across several methods. 
We attribute this to the centered nature of targets in RGBNT201, where minor shifts effectively reduce background clutter and act as implicit data augmentation. 
In severe occlusion cases (mask area ratios of 0.3 and 0.4), IDEA~\cite{wang2025idea} achieves marginally better mAP than PRISM, likely due to its integration of textual semantic cues that complement visual features under extreme information loss. 
This further supports our central thesis that enriching representations with auxiliary semantic information significantly enhances model resilience in adverse conditions.

\begin{table}[t]
  \caption{ Performance comparison of PFN with varying component configurations on RGBNT201}
  \label{tab:PFN_ablation2}
  \centering
  \renewcommand\arraystretch{1.0}
  \setlength{\tabcolsep}{11.75pt}
  \begin{tabular}{cccccc}
      \noalign{\hrule height 1pt}
      \multicolumn{1}{c}{\multirow{2}{*}{\textbf{Index}}} &\multicolumn{3}{c}{\textbf{PFN}} & \multicolumn{2}{c}{\textbf{Metrics}} \\
      \cmidrule(r){2-4} \cmidrule(r){5-6}
& \textbf{Intra}              & \textbf{Inter}                & \textbf{Tri}                   & \textbf{mAP}    & \textbf{Rank-1}   \\\hline
A                   & 1                  & 1                 & 1 & 77.2  & 81.0   \\
B                  & 1                 & 1                  & 2                    & 76.3  & 80.6                    \\
C                  & 1                 & 1                 & 3                     & 78.6  & 83.5                \\
D                   & 1                  & 2                 & 1 & 79.4  & 81.8   \\
\rowcolor[gray]{0.92}
E                  & 1                 & 2                  & 2                    & 80.5  & 84.0                    \\
F                  & 1                 & 2                 & 3                     & 80.2  & 84.8                \\
G                   & 1                  & 3                 & 1 & 76.8  & 81.1   \\
H                  & 1                 & 3                  & 2                    & 78.9  & 83.9                    \\
I                  & 1                 & 3                 & 3                     & 78.8  & 84.2                \\
  \noalign{\hrule height 1pt}
  \end{tabular}
\end{table}

\begin{table}[t]
  \caption{Comparison of computational complexity between PS6 and standard attention modules under varying sequence lengths. Results confirm the linear complexity of PS6, in contrast to the quadratic growth of self- and cross-attention}
  \label{tab:seqlen_supp}
  \centering
  \renewcommand\arraystretch{1.18}
  \setlength{\tabcolsep}{20.65pt}
  \begin{tabular}{r|cc}
      \noalign{\hrule height 1pt}
      \textbf{SeqLen} & PS6 & SelfAttn/CrossAttn \\\hline
      64 & 0.0840 & 0.2139\\
      128 & 0.1679 & 0.4530\\
    256 & 0.3358	&1.0065\\
      512 & 0.6716	&2.4159\\
      1024 & 1.3432	&6.4425\\
      2048 & 2.6865	&19.3275\\
      \noalign{\hrule height 1pt}
  \end{tabular}
\end{table}

\begin{table}[t]
  \caption{Inference time comparison of different methods}
  \label{tab:inf}
  \centering
  \renewcommand\arraystretch{1.18}
  \setlength{\tabcolsep}{1.9pt}
  \begin{tabular}{r|ccccc}
      \noalign{\hrule height 1pt}
      \textbf{Methods} & \textbf{PromptMA} &\textbf{MambaPro} & \textbf{IDEA} & \textbf{MFRNet} & \textbf{Ours} \\
      \hline
      \textbf{Total Inf. Time (s)} & 9.75 & 15.90 & 21.88 &10.77 & 11.78 \\
      \textbf{Avg. Batch Time (ms)} & 361.11 & 590.00 &810.48  & 399.02 & 436.34  \\
      \noalign{\hrule height 1pt}
  \end{tabular}
\end{table}

\begin{table}[t]
  \caption{Efficiency comparison with SOTA methods in terms of computational cost, inference speed, and memory usage. Our method achieves the lowest memory consumption while maintaining competitive throughput and FLOPs}
  \label{tab:gflop_supp}
  \centering
  \renewcommand\arraystretch{1.18}
  \setlength{\tabcolsep}{4.5pt}
  \begin{tabular}{r|ccc}
      \noalign{\hrule height 1pt}
       \textbf{Methods} & \textbf{GFLOPs $\downarrow$} & \textbf{Sample/s $\uparrow$}  &\textbf{Memory (MiB) $\downarrow$} \\\hline
      PromptMA~\cite{zhang2025prompt} & 67.40 &119.5 &3647 \\
      MambaPro~\cite{wang2025mambapro}	&51.25	&106.7 &2761 \\
      IDEA~\cite{wang2025idea}	&43.73	&117.2 &2760 \\
      MFRNet w/ ViT-S~\cite{feng2025multi} & 22.10 & 155.8 &3404 \\
      Ours	&40.59	&112.7 &1718 \\
      \noalign{\hrule height 1pt}
  \end{tabular}
\end{table}

\begin{table}[t]
  \caption{Performance comparison of PRISM with varying component configurations on MSVR310}
  \centering
  \renewcommand\arraystretch{1.0}
    \setlength{\tabcolsep}{10.75pt}
  {
  \begin{tabular}{cccccc}
      \noalign{\hrule height 1pt}
      \multicolumn{1}{c}{\multirow{2}{*}{\textbf{Index}}} &\multicolumn{3}{c}{\textbf{Modules}} & \multicolumn{2}{c}{\textbf{Metrics}} \\
      \cmidrule(r){2-4} \cmidrule(r){5-6}
 & \textbf{Mask}              & \textbf{SDTP}                & \textbf{PFN}                   & \textbf{mAP}    & \textbf{Rank-1}   \\\hline
  A                  & -                  & \XSolidBrush                  & \XSolidBrush                    & 40.8  & 55.7 \\
  B                  & -                  & \XSolidBrush                  & \CheckmarkBold                      & 44.0  & 58.4 \\
  \multirow{1}{*}{C} & \multirow{1}{*}{\XSolidBrush} & \multirow{1}{*}{\CheckmarkBold} & \multirow{1}{*}{\CheckmarkBold}    & 46.2  & 61.8 \\
  \rowcolor[gray]{0.92}
  \multirow{1}{*}{D} & \multirow{1}{*}{\CheckmarkBold} & \multirow{1}{*}{\CheckmarkBold} & \multirow{1}{*}{\CheckmarkBold}    &\textbf{47.6} &\textbf{64.8}  \\
  \noalign{\hrule height 1pt}
  \end{tabular}
  } 
  \label{tab:sup310}
\end{table}

\begin{table}[t]
  \caption{Quantitative Evaluation of Cosine Similarity Distributions Using Separation and Overlap Metrics, with mAP Comparison}
  \label{tab:cos}
  \centering
  \renewcommand\arraystretch{1.18}
  \setlength{\tabcolsep}{8pt}
  \begin{tabular}{r|ccc}
      \noalign{\hrule height 1pt}
      \textbf{Metric} & \textbf{Separation Metric} & \textbf{Overlap Area} & \textbf{mAP} \\
      \hline
       \textbf{MambaPro}~\cite{wang2025mambapro} &3.5274 & 0.0872 & 78.9  \\
      \textbf{IDEA}~\cite{wang2025idea} & 2.9998 & 0.0882 & 80.2  \\
      \textbf{PRISM (Ours)} & 3.5959 &0.0659  & 80.5 \\
      \noalign{\hrule height 1pt}
  \end{tabular}
\end{table}

\subsection{Ablation Studies}\label{sec:ablation}
We evaluate the effectiveness of key modules on RGBNT201, using a baseline that leverages only the class tokens from the visual encoder. 
We set the number of stacked layers in TAFE to $k = 2$ for optimal performance, with the stacking architecture illustrated in Fig.~\ref{framework}.  
Due to the lightweight design of each module, this configuration introduces only a minimal number of additional parameters.

\subsubsection{Effects of Key Modules}
\label{subsec:ablation_key_modules}
Table~\ref{tab:main_ablation} presents the performance of various combinations of the proposed components. 
Model A serves as the baseline, achieving an mAP of 70.5\% and a Rank-1 accuracy of 73.4\%. 
Incorporating PFN into the framework, Model B improves the performance to an mAP of 73.7\% and a Rank-1 accuracy of 78.1\%, demonstrating the benefit of this integration. 
Model C further integrates SDTP, boosting the mAP to 75.8\% and the Rank-1 accuracy to 79.3\%. 
Finally, Model D introduces semantic details through the use of a mask extractor, achieving the highest performance with an mAP of 80.5\%. 
These results confirm the effectiveness of the proposed components in improving overall performance.

\subsubsection{Effects of Key Components in SDTP}
\label{subsec:ablation_sdtp}
Table~\ref{tab:SDTP_ablation} evaluates the impact of different components within SDTP. 
Model A (baseline, without SDTP) achieves an mAP of 73.7\% and a Rank-1 accuracy of 78.1\%. 
Model B adds the TAFE module, increasing the mAP to 76.4\%, which shows the positive effect of fine-grained semantic interaction. 
Models C, D, and E include the TDM module with varying numbers of PS6 modules: Model C (1 PS6) achieves 77.5\% mAP, Model D (3 PS6s) reaches 79.3\% mAP, and Model E (2 PS6s) achieves the highest mAP of 80.5\%. 
These results confirm that TDM contributes significantly to performance gain, with two PS6 modules yielding the optimal configuration.

\subsubsection{Effects of Key Components in PFN}
\label{subsec:ablation_pfn}
Table~\ref{tab:PFN_ablation} shows the performance of different components within PFN. 
Model A (without PFN) achieves an mAP of 76.3\% and a Rank-1 accuracy of 79.9\%. 
Model B incorporates IM, improving the mAP to 76.4\% and the Rank-1 accuracy to 81.3\%. 
Model C further integrates IT, boosting the mAP to 77.6\%, which demonstrates the effectiveness of PS6-based interaction. 
Finally, Model D introduces TA, achieving the best performance with an mAP of 80.5\%. 
The results validate the contribution of each component in the PFN framework, particularly the progressive three-stage aggregation strategy.

\subsubsection{Effects of Different Mask Generation Methods}
We evaluate the impact of mask generation strategies on RGBNT201 and RGBNT100, as shown in Tables~\ref{tab:data}, \ref{tab:mask4100}, and \ref{tab:inf_supp}.
On RGBNT201 (person ReID), OpenPifPaf outperforms SAM2 because pose keypoints provide strong structural priors for articulated bodies, yielding more accurate masks that localize key parts. 
Conversely, on RGBNT100 (vehicle ReID), SAM2 with box prompts achieves better results, as rigid objects are poorly represented by sparse keypoints. 
In this case, OpenPifPaf often produces fragmented or incomplete masks (Fig.~\ref{fig:mask}), which degrade performance by introducing noise rather than guidance.

These results confirm that the optimal mask generator depends on the target object’s physical structure. 
To avoid severe failures, such as missing the object entirely, we adopt an adaptive prompting strategy that uses skeleton-based masking for pedestrians and box or center-prior masking for vehicles. 

Importantly, all masks are generated from the RGB modality only. This choice is empirically justified for two reasons. First, RGB provides the most reliable visual cues for pre-trained segmentation models. Second, our framework exhibits robustness to minor mask imperfections or inter-modal misalignments since the mask serves as a coarse semantic prior rather than pixel-accurate supervision. This robustness is further enhanced by the Cross-Modal Union (CMU) mechanism in the Tail Drop Module, which aggregates token decisions across modalities. Consequently, using a single RGB-based generator avoids unnecessary complexity while maintaining effectiveness, as verified by the stability under pixel offsets and occlusions in Table~\ref{tab:pixel_offset} and Table~\ref{tab:occluded_area}.

\subsubsection{Effects of Drop Rate in the TDM}
Table~\ref{tab:thyper} evaluates the performance of PRISM under different drop rates in the TDM, with experiments conducted on RGBNT201.
The results show that the CMU-based union strategy performs better than modality-specific tail dropping. 
A drop rate of 0.5 achieves the best performance.
Performance degrades when the drop rate deviates from this value, indicating that 0.5 strikes an effective balance between pruning poorly attended tokens and preserving informative ones.

Notably, the actual set of dropped tokens is determined by the intersection of the drop regions across the three modalities, resulting in an effective dropout ratio that is inherently lower than the configured value. 
Therefore, setting the drop rate to 0.5 ensures sufficient sparsification without compromising cross-modal alignment.
This choice aligns with the well-established dropout principle~\cite{krizhevsky2012imagenet}, where a 50\% retention rate maximizes subnetwork diversity and enhances ensemble effects during training.

\subsubsection{Ablation Study on Number of Modules in PFN}
To determine the optimal stacking depth for the IT and TA modules within PFN, we conduct an ablation study on their architectural depth, as shown in Table~\ref{tab:PFN_ablation2}. 
The results indicate that setting the number of stacked modules to $k = 2$ yields the best performance. 
Further increasing the depth provides no additional gain, suggesting diminishing returns beyond this point.
The lack of improvement with deeper stacking may stem from redundant feature refinement or increased optimization difficulty in very deep fusion pathways. 
Notably, each module is designed to be lightweight with few trainable parameters; thus, doubling the stack count results in only a marginal increase in overall model complexity, which highlights the parameter efficiency of our design. 
Nevertheless, excessive depth may impair generalization, particularly on the RGBNT201 dataset, where challenging imaging conditions and suboptimal image quality place high demands on model robustness.

\begin{figure*}[t]
  \centering
    {
  \includegraphics[width=\linewidth]{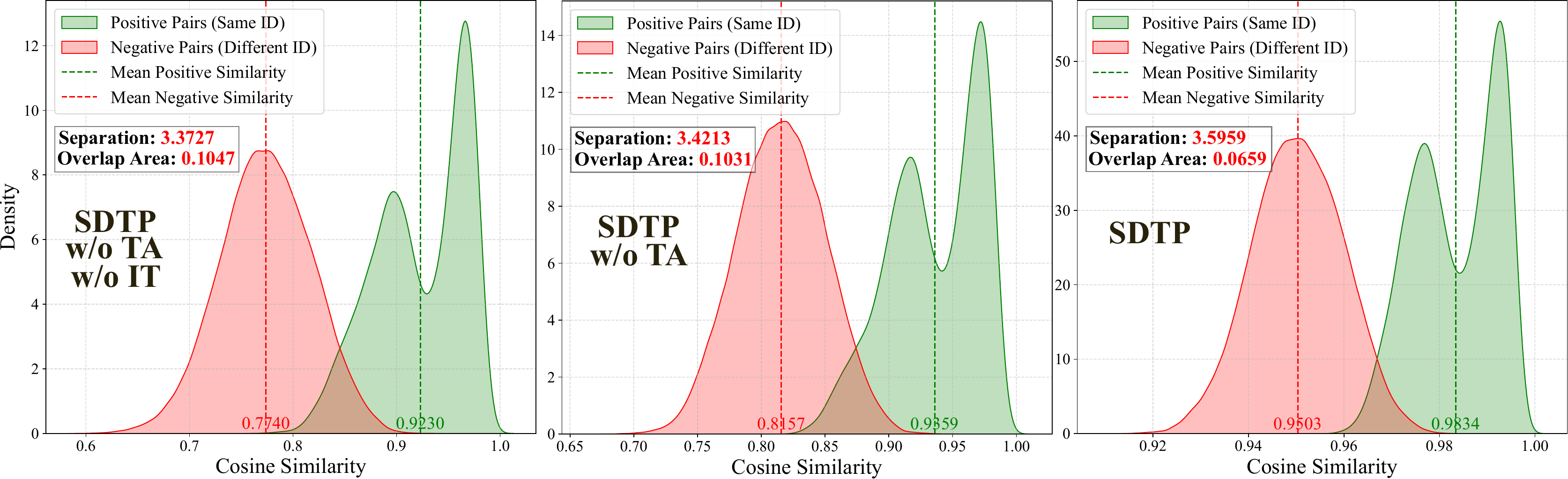}
  }
   \vspace{-4mm}
   \caption{Cosine similarity distribution visualization. Incorporating IT and TA improves separation and reduces overlap, indicating enhanced discriminative power from progressive interaction.}
  \label{fig:cosine}
\end{figure*}

\begin{figure*}[t]
  \centering
    \resizebox{1.0\textwidth}{!}
    {
  \includegraphics[width=1\linewidth]{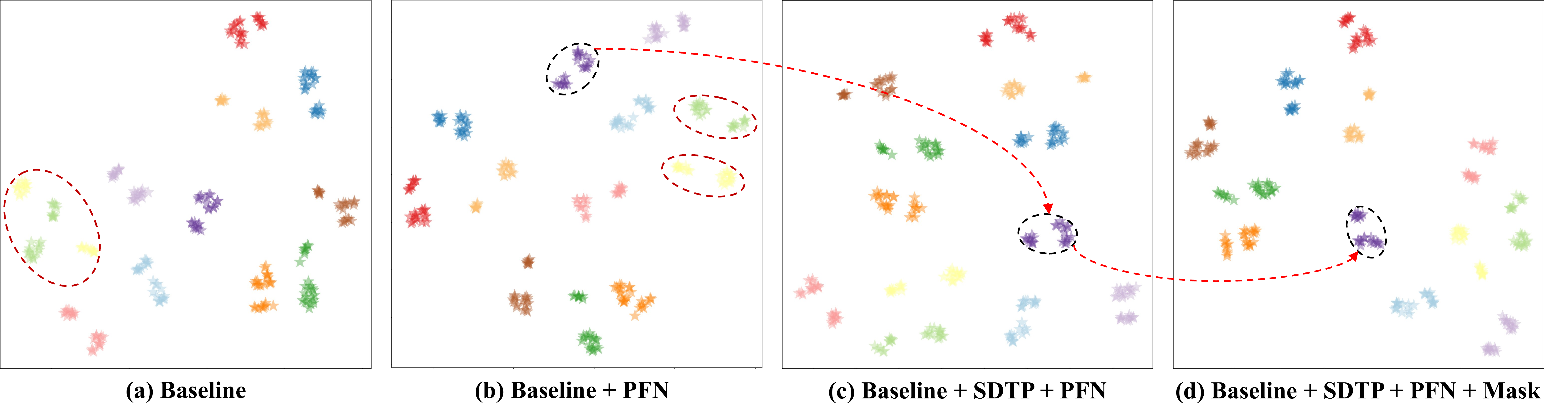}
  }
   \caption{Visualization of feature distributions using t-SNE~\cite{van2008visualizing}. Different colors represent different identities, illustrating the clustering effect and separability achieved by our method.}
  \label{fig:tsne}
  \vspace{-5mm}
\end{figure*}

\subsubsection{Inference Efficiency}
The proposed SDTP and PFN modules are built upon the lightweight PS6 backbone and introduce only marginal computational overhead, as illustrated in Fig.~\ref{fig1}(b). 
To evaluate the efficiency of PS6, we compare its GFLOPs with those of a standard self-attention module under identical input conditions, as shown in Table~\ref{tab:seqlen_supp}. 
The results confirm that PS6 scales linearly with sequence length, in contrast to the quadratic complexity of standard self-attention.

We compare inference time, GFLOPs, sampling speed, and memory consumption with recent SOTA methods in Tables~\ref{tab:inf} and~\ref{tab:gflop_supp} to assess deployment efficiency. All FLOPs are measured using fvcore under identical experimental settings.
As shown, our method achieves comparable inference speed to PromptMA and MFRNet, while significantly outperforming MambaPro and IDEA by a large margin.
Crucially, under consistent experimental settings, our approach attains the lowest memory consumption among all compared methods. 
Moreover, its memory usage is substantially lower than that of PromptMA and MFRNet, while still maintaining competitive throughput and FLOPs, which highlights its advantage for practical deployment.

This efficiency stems from the PS6 mechanism, which minimizes the footprint of intermediate activations. 
In contrast, MambaPro processes each modality with independent Mamba blocks, requiring multiple parallel state spaces during inference. 
Although MFRNet adopts a lighter ViT-S backbone (compared to the ViT-B used by other methods), its memory consumption remains high due to the Mixture-of-Experts (MoE) based fusion module. Specifically, only a subset of experts is activated per token, keeping FLOPs low and enabling marginally faster inference. However, all expert parameters and routing logits must still be retained in memory. Coupled with dense feature fusion operations that produce large intermediate activation tensors, this results in a substantial memory footprint despite its lower computational cost.
Moreover, PromptMA achieves marginally faster inference than our method, primarily owing to its fixed-token processing pipeline that avoids dynamic token selection. However, by operating on the full set of input tokens throughout the network without explicit token reduction, it incurs higher FLOPs and greater memory usage, particularly in intermediate feature maps. In contrast, PRISM’s semantic-aware token pruning effectively reduces sequence length early in the network, yielding lower computational cost and a reduced memory footprint with minimal latency overhead.

\subsubsection{Effect of Key Modules on a Degraded Vehicle Dataset}
\label{subsec:ablation_vehicle}
As shown in Table~\ref{tab:sup310}, we further evaluate the proposed framework on MSVR310, a challenging vehicle ReID dataset with significant real-world degradations, including low resolution, motion blur, and adverse weather. 
The baseline achieves 40.8\% mAP and 55.7\% Rank-1 accuracy under these conditions. 
Integrating the Progressive Fusion Network (PFN) improves performance to 44.0\% mAP and 58.4\% Rank-1, demonstrating its robustness in fusing multi-spectral features under degraded inputs. 
Further incorporating the SDTP module with semantic mask guidance achieves the best performance of 47.6\% mAP and 64.8\% Rank-1.

The SDTP module is explicitly designed to leverage semantic structure as a prior for reliable region identification and token pruning. 
When SDTP operates without this guidance, it relies solely on response-driven region selection, where high-activation areas in the feature map are prioritized as potentially discriminative.
This contrast highlights the effectiveness of our semantic-aware token pruning mechanism, where the synergy between semantic structure and multi-spectral features is key to robust matching.
Overall, these results validate the complementary design of the proposed components and their ability to handle complex, real-world vehicle ReID scenarios.

\begin{figure*}[t]
    \centering
     \resizebox{1.0\textwidth}{!}
    {
    \includegraphics[width=30\linewidth]{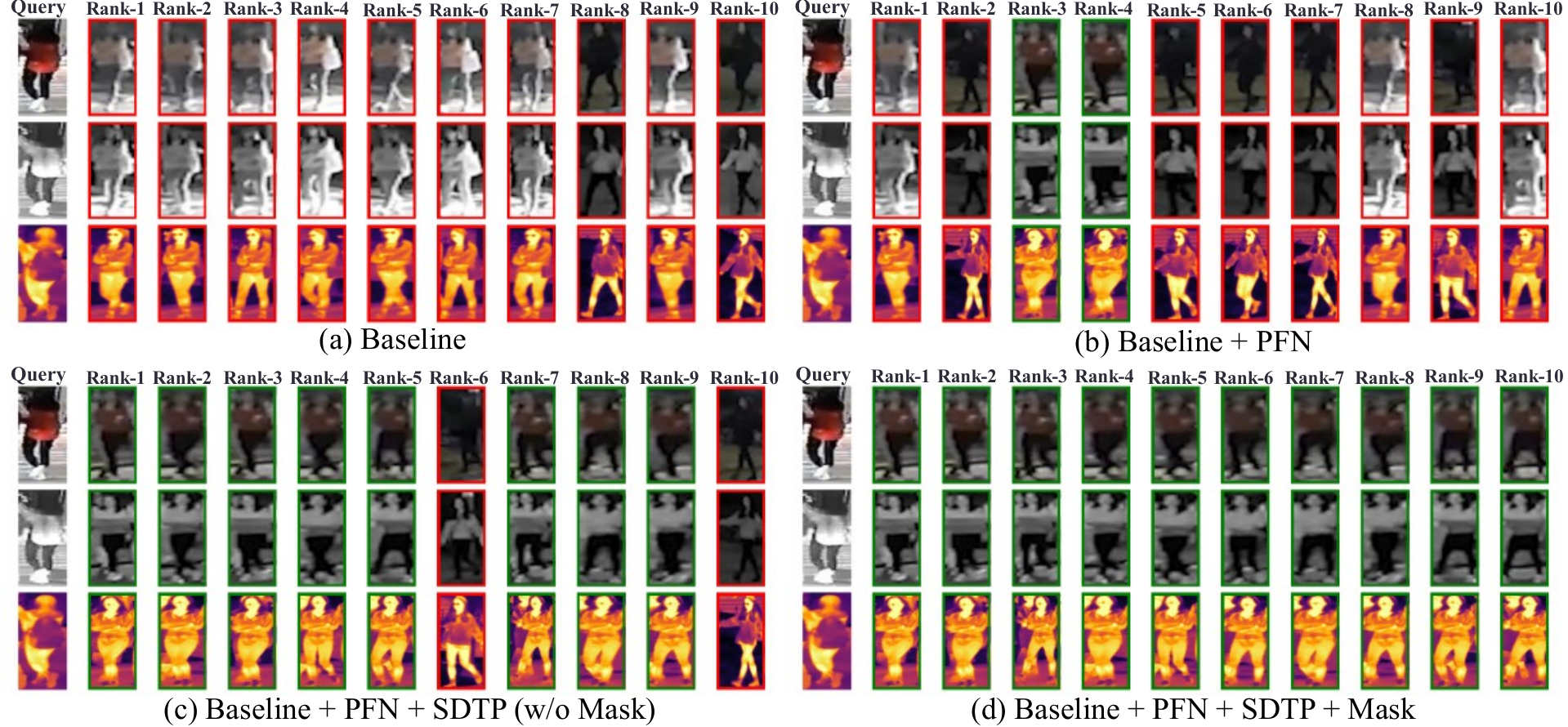}
    }
     \caption{Rank list comparison of our model with varying components on RGBNT201. 
The progressive performance improvement with the addition of each module demonstrates the effectiveness of the proposed architecture.}
    \label{fig:rank_sup}
\end{figure*}

\begin{figure*}[t]
    \centering
     \resizebox{1.0\textwidth}{!}
    {
    \includegraphics[width=30\linewidth]{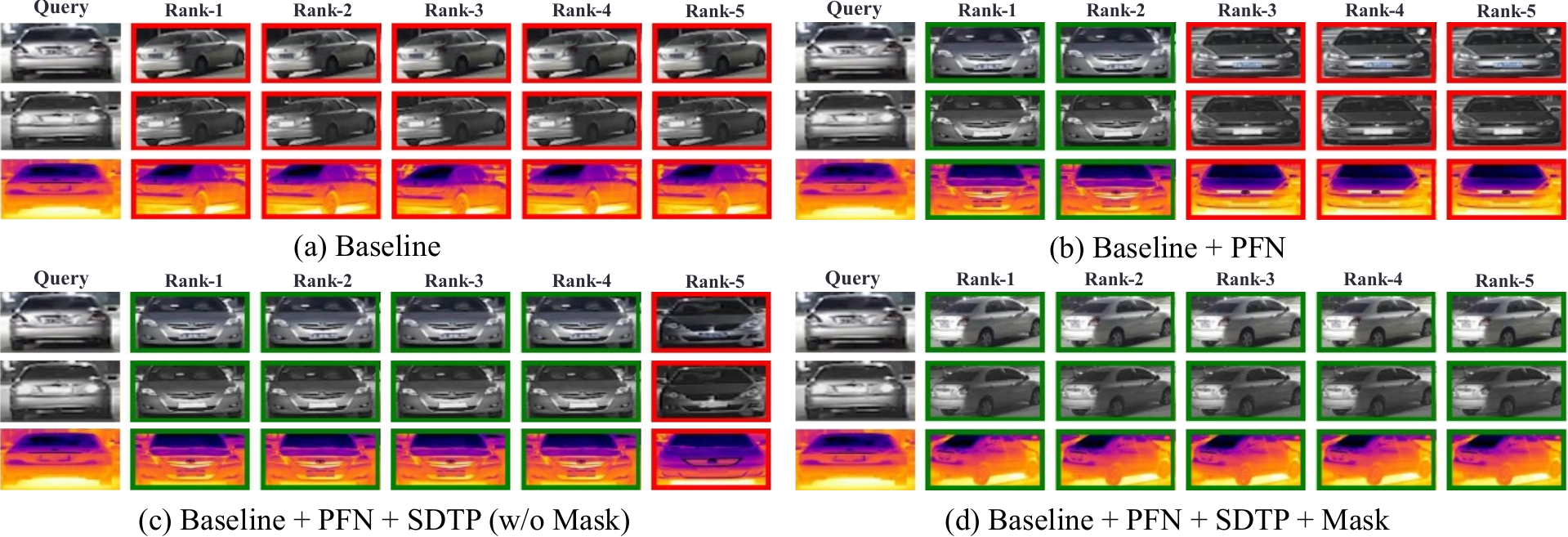}
    }
     \caption{Rank list comparison of our model with varying components on RGBNT100.}
    \label{fig:rank_sup_car}
\end{figure*}

\begin{table*}[t]
  \caption{Performance of PS6 in Image Fusion}
  \label{tab:tcmoa}
  \centering
  \renewcommand\arraystretch{1.18}
  \setlength{\tabcolsep}{18.65pt}
  \begin{tabular}{r|ccccccc}
      \noalign{\hrule height 1pt}
      \textbf{Model} & EN $\uparrow$ & SD $\uparrow$ & VIF $\uparrow$ & AG $\uparrow$ & CC $\uparrow$ & SCD $\uparrow$ &MS\_SSIM $\uparrow$ \\\hline
      Baseline~\cite{zhu2024task} & 7.428 & 9.805 & 0.726 &4.659	&0.673	&1.546 &0.949 \\
      Baseline w/ PS6 & 7.435 & 10.624 & 0.735 &4.757	&0.676	&1.593 &0.959\\
      \noalign{\hrule height 1pt}
  \end{tabular}
\end{table*}

\begin{figure}[t]
    \centering
    {
    \includegraphics[width=\linewidth]{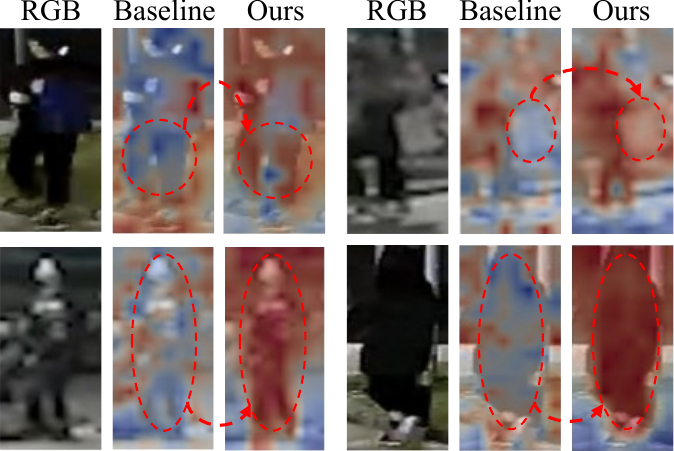}
    }
    \vspace{-4mm}
    \caption{Visualization of channel activation maps. Interacting with semantic masks helps the model focus on more discriminative regions, thereby boosting feature robustness and enhancing interpretability.}
  \label{fig:map}
\end{figure}

\begin{figure}[t]
    \centering
    {
    \includegraphics[width=\linewidth]{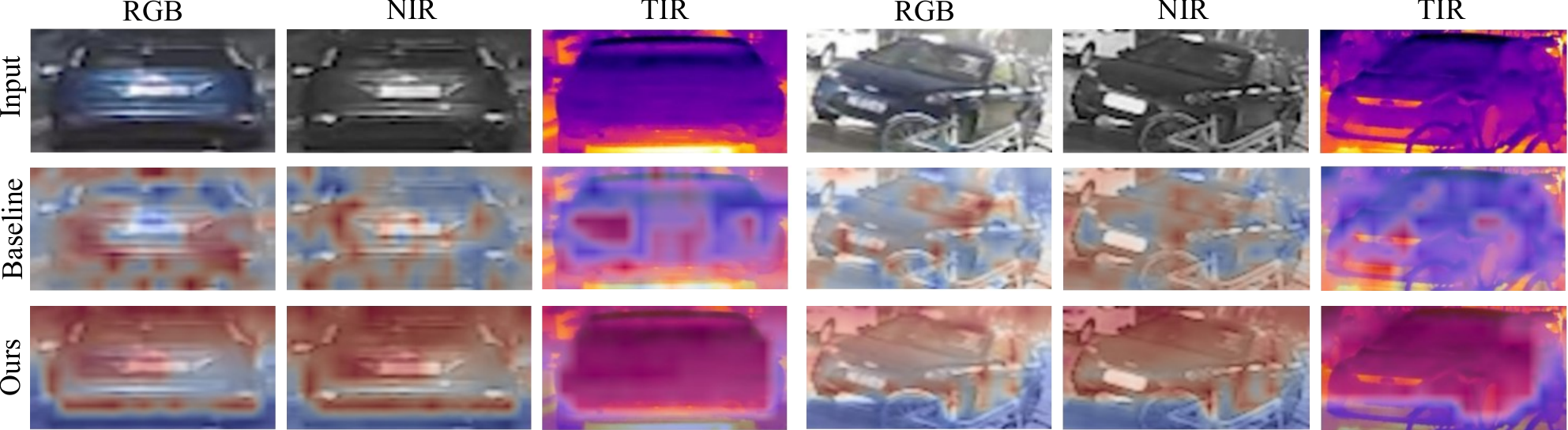}
    }
    \vspace{-4mm}
    \caption{Visualization of channel activation maps for vehicle ReID. The interaction with semantic masks enables the model to focus on more discriminative regions, thereby improving feature robustness and enhancing interpretability.}
  \label{fig:map_car}
  \vspace{-4mm}
\end{figure}

\subsection{Visualization Analysis}

\subsubsection{Cosine Similarity Distributions}
\label{sec:cos}
Fig.~\ref{fig:cosine} shows the distributions of cosine similarities for features after adding different components to PFN.
Unlike methods such as IDEA~\cite{wang2025idea}, which typically only examine the separation (e.g., peak difference) between the two distribution modes, our approach focuses on evaluating the comprehensive overlap area and separation degree of the entire distributions. To quantify this, we utilize two specific metrics. The Separation Metric is defined as the standardized mean difference between the positive samples ($X_1$) and negative samples ($X_2$). This is calculated as follows:
\begin{equation}
\text{Separation Metric} = (\bar{X}_1 - \bar{X}_2) / \sqrt{(s_1^2 + s_2^2) / 2},
\end{equation}
where $\bar{X}_1$ and $\bar{X}_2$ are the means of the positive and negative samples, and $s_1^2$ and $s_2^2$ are their respective variances. Furthermore, to estimate the overlap between the distributions, we use Kernel Density Estimation (KDE). The overlap area is approximated by the integral of the minimum of the two probability density functions:
\begin{equation}
\text{Overlap Area} \approx \int \min(f_1(x), f_2(x)) dx,
\end{equation}
where $f_1(x)$ and $f_2(x)$ are the probability density functions estimated via KDE for the positive and negative samples, respectively. Together, these metrics provide a concise framework for evaluating both the separation and similarity of the two distributions, as shown in Fig.~\ref{fig:cosine}.
The results show that both the IT and TA modules independently improve the separation metric and reduce the overlap area. 
When combined, they achieve optimal performance with a separation metric of 3.5959 and an overlap area of 0.0659, demonstrating their effectiveness in enhancing feature discriminability.
Furthermore, as shown in Table~\ref{tab:cos}, our method outperforms SOTA approaches in both distribution quality and retrieval accuracy.

\subsubsection{Multi-Modal Feature Distributions}
\label{subsec:tsne_analysis}

Fig.~\ref{fig:tsne} visualizes the learned feature embeddings from different configurations.
As shown in Fig.~\ref{fig:tsne}(a), the baseline model exhibits entangled clusters with significant intra-class variance, particularly in the region highlighted by the red circle. 
In contrast, the integration of the Progressive Fusion Network (PFN) yields better-separated class boundaries and reduced intra-class dispersion (Fig.~\ref{fig:tsne}(b)). This demonstrates that PFN effectively fuses complementary multi-spectral cues to enhance discriminability.
Further improvements are observed when incorporating the SDTP module with semantic guidance, as shown in Fig.~\ref{fig:tsne}(c) and (d). 
Compared to the features with PFN only, the resulting features exhibit tighter intra-class clustering and clearer inter-class separation. 
This suggests that SDTP, by leveraging semantic structure to guide token pruning and feature enhancement, suppresses irrelevant background responses.

\subsubsection{Rank List Comparison}
Fig.~\ref{fig:rank_sup} and Fig.~\ref{fig:rank_sup_car} compare the cross-camera rank lists produced by the baseline method and progressively enhanced variants, culminating in the full PRISM model.
Starting from the baseline, which exhibits noisy and inconsistent ranking patterns, the introduction of the tri-modal fusion module leads to an improved retrieval order.
Further incorporating the semantic-guided token selection mechanism enhances ranking stability.
This visual ablation clearly demonstrates the individual and cumulative contributions of the proposed components, highlighting PRISM's ability to retrieve correct matches across camera views.

\subsubsection{Visualization of Channel Activation Maps}
Fig.~\ref{fig:map} and Fig.~\ref{fig:map_car} illustrate a comparison between the channel activation maps of our baseline model and those generated by PRISM following the TAFE module.
The incorporation of PS6-based fine-grained semantic interaction enables the model to focus on more discriminative regions, thereby enhancing feature robustness and interpretability.

\subsection{Generalization Capability of PS6 in Multi-Modal Tasks}
Unlike recent works that use global semantic information for feature sampling or guidance, our framework incorporates semantic priors as full inputs to enable finer-grained local interaction. Moreover, the proposed PS6 module achieves tri-modal feature interaction with linear computational complexity and fewer parameters, making it highly suitable for resource-constrained multi-modal applications.

As shown in Table~\ref{tab:tcmoa}, integrating PS6 into a baseline fusion network improves performance across multiple metrics, demonstrating its effectiveness in capturing cross-modal correspondences. 
The gains in EN, SD, and AG indicate that the fused features preserve richer spatial details and sharper edges, while improvements in CC, SCD, and MS\_SSIM suggest better structural alignment with the reference image. 
These results confirm that PS6 enhances both detail preservation and cross-modal consistency, and highlight its broader applicability to low-level multi-modal vision tasks such as image fusion.

\section{Conclusion}
\label{sec:conclusion}
In this paper, we propose PRISM, a novel multi-modal object Re-Identification framework built upon Prompt-S6 (PS6) and semantic-aware knowledge guidance. 
Specifically, we first utilize a pre-trained mask generator to obtain semantic masks, providing spatially aligned guidance for foreground-background separation. 
Building upon PS6, our Semantic-Driven Token Pruning (SDTP) facilitates fine-grained interaction between semantic information and tri-modal features while suppressing background-associated tokens through a cross-modal consensus mechanism.
Furthermore, our Progressive Fusion Network (PFN) progressively models intra-modal dynamics, inter-modal interactions, and tri-modal alignment, achieving comprehensive aggregation across RGB, NIR, and TIR modalities.
Benefiting from the linear complexity of PS6, both SDTP and PFN achieve efficient cross-modal interaction with minimal computational overhead.
Extensive experiments on four public multi-modal object ReID benchmarks confirm the effectiveness and efficiency of our approach.

\bibliographystyle{IEEEtran}
\bibliography{Ref}

\vfill
\end{document}